\documentclass[iicol,sn-mathphys-num]{sn-jnl}

\usepackage{graphicx}%
\usepackage{multirow}%
\usepackage{amsmath,amssymb,amsfonts}%
\usepackage{amsthm}%
\usepackage{mathrsfs}%
\usepackage[title]{appendix}%
\usepackage{xcolor}%
\usepackage{textcomp}%
\usepackage{manyfoot}%
\usepackage{booktabs}%
\usepackage{algorithm}%
\usepackage{algorithmicx}%
\usepackage{algpseudocode}%
\usepackage{listings}%

\usepackage{url}
\usepackage{xspace}
\usepackage{hyperref}
\usepackage{cleveref}

\def\eg{\emph{e.g.}} 
\def\ie{\emph{i.e.}} 
 
\def\etc{\emph{etc.}} \def\vs{\emph{vs.}}

\definecolor{newcolor}{rgb}{.8,.349,.1}

\newcommand{\xie}[1]{{\color{black} #1}}
\newcommand{\xienew}[1]{{\color{black} #1}}
\def\qi{\textcolor{black}}
\def\qinew{\textcolor{black}}

\theoremstyle{thmstyleone}%
\theoremstyle{thmstyletwo}%

\theoremstyle{thmstylethree}%

\begin{document}

\title[MVL Survey]{A Survey of Medical Vision-and-Language Applications \\ and Their Techniques}


\author[1]{\fnm{Qi} \sur{Chen}}\equalcont{These authors contributed equally to this work.}

\author[2]{\fnm{Ruoshan} \sur{Zhao}}\equalcont{These authors contributed equally to this work.}

\author[1]{\fnm{Sinuo} \sur{Wang}}

\author[1]{\fnm{Vu Minh Hieu} \sur{Phan}}

\author[1]{\fnm{Anton van den} \sur{Hengel}}

\author[1]{\fnm{Johan} \sur{Verjans}}

\author[1]{\fnm{Zhibin} \sur{Liao}}

\author[3]{\fnm{Minh-Son} \sur{To}}

\author[4]{\fnm{Yong} \sur{Xia}}

\author[2]{\fnm{Jian} \sur{Chen}}

\author*[1]{\fnm{Yutong} \sur{Xie}}\email{yutong.xie678@gmail.com}

\author[1]{\fnm{Qi} \sur{Wu}}

\affil[1]{\orgdiv{Australian Institute for Machine Learning}, \orgname{University of Adelaide}, \orgaddress{\country{Australia}}}

\affil[2]{\orgdiv{School of Software Engineering}, \orgname{South China University of Technology}, \orgaddress{\country{China}}}

\affil[3]{\orgname{South Australia Medical Imaging}, \orgaddress{\country{Australia}}}

\affil[4]{\orgdiv{School of Computer Science \& Engineering}, \orgname{Northwestern Polytechnical University}, \orgaddress{\country{China}}}

\abstract{
Medical vision-and-language models (MVLMs) have attracted substantial interest due to their capability to offer a natural language interface for interpreting complex medical data. Their applications are versatile and have the potential to improve diagnostic accuracy and decision-making for individual patients while also contributing to enhanced public health monitoring, disease surveillance, and policy-making through more efficient analysis of large data sets. MVLMS integrate natural language processing with medical images to enable a more comprehensive and contextual understanding of medical images alongside their corresponding textual information. Unlike general vision-and-language models trained on diverse, non-specialized datasets, MVLMs are purpose-built for the medical domain,
automatically extracting and interpreting critical information from medical images and textual reports to support clinical decision-making.
Popular clinical applications of MVLMs include automated medical report generation, medical visual question answering, medical multimodal segmentation, diagnosis and prognosis and medical image-text retrieval. Here, we provide a comprehensive overview of MVLMs and the various medical tasks to which they have been applied.
We conduct a detailed analysis of various vision-and-language model architectures, focusing on their distinct strategies for cross-modal integration/exploitation of medical visual and textual features. We also examine the datasets used for these tasks and compare the performance of different models based on standardized evaluation metrics.
Furthermore, we highlight potential challenges and summarize future research trends and directions. The full collection of papers and codes is available at: \href{https://github.com/YtongXie/Medical-Vision-and-Language-Tasks-and-Methodologies-A-Survey}{https://github.com/MVLMs}.}

\keywords{Vision and language, Medical applications, Survey}



\maketitle

\section{Introduction}


\begin{figure*}[h!]
    \centering
    \includegraphics[width=1.0\textwidth]{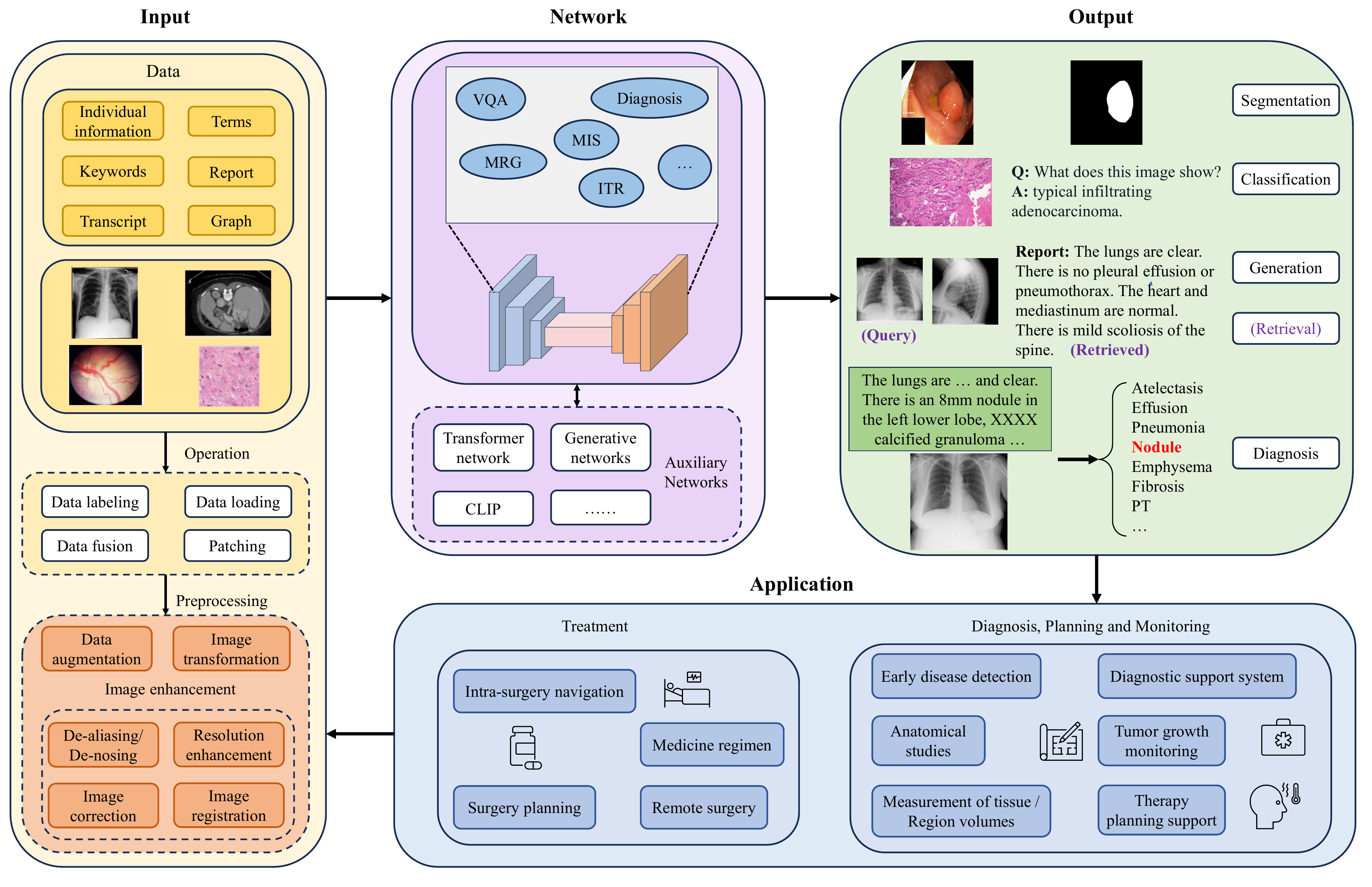}
    \caption{This diagram \qinew{illustrates the components of MVLMs, including data inputs like medical images, reports, and graphs, along with preprocessing operations such as data fusion and enhancement. It leverages various model architectures (\eg, transformer and generative networks) for tasks like medical report generation (MRG), visual question answering (VQA), medical image segmentation (MIS), and image-text retrieval (ITR). Applications span diagnosis, surgery planning, and early disease detection, enhancing clinical decision-making and workflows.}
    }
    \label{framework}
\end{figure*}
\begin{figure*}
    \centering
    \includegraphics[width=1.0\textwidth]{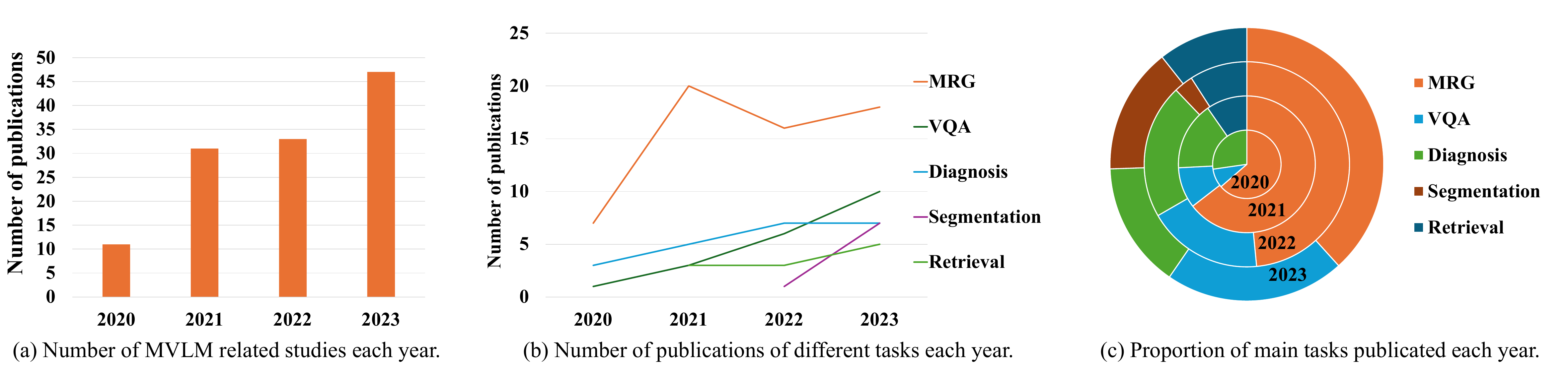}
    \caption{Statistics on the number of papers published in several top journals and conferences such as IEEE-TMI, Medical Image Analysis, CVPR, ICCV, MICCAI, \etc~The plot shows consistent growth in recent literature.}
    \label{fig:num_papers}
\end{figure*}

\qinew{The exponential growth of medical data, particularly multimodal data in recent years, has heightened the need for medical vision-language models (MVLMs) that integrate computer vision and natural language processing, leveraging complementary features from the data to improve planning, prediction, diagnosis, and treatment in medical practice.}
MVLMs often learn visual features through image encoders and textual features through text encoders, and then follow up with specific generators or classifiers devised for individual tasks. They can be trained to have powerful data comprehension and generation capabilities that can be used in various medical applications \cite{shamshad2023transformers}. For example, they can analyze image and text data, and then generate diagnostic reports \cite{li2023dynamic,chen2020generating}, provide diagnostic recommendations \cite{cui2023deep,song2022multicenter,liu2023multi,lu2022m} or enable efficient retrieval of medical images and text data~\cite{abacha20233d} to facilitate clinical research and case analysis, as shown in Figure \ref{framework}.  MVLMs can reduce the workload of doctors and improve the accuracy of diagnosis and treatment, and have thus become an important topic in the field of medical image analysis. A visualization of the recent growth in the field is provided in Figure \ref{fig:num_papers}.

\xie{MVLMs are specifically designed to handle complex and specialized medical data, such as medical images and electronic medical records, all of which require an understanding of domain-specific terminology, subtle visual patterns, and contextually relevant clinical knowledge. The demands on MVLMs are higher as their outputs can directly impact patient care, requiring both high precision and reliability. 
\qinew{However, developing large vision-language models in the medical domain poses several challenges.}
%
\qinew{First, collecting medical data is difficult due to privacy concerns and limited accessibility.}
Stringent regulations often restrict medical data sharing, limiting the availability of large, annotated datasets necessary for training.
Second, medical data is highly heterogeneous, involving various imaging modalities (\eg, X-ray, MRI, and CT) and diverse documentation styles, which complicates the development of models that can generalize effectively. There is also a critical need for domain expertise to ensure high-quality data annotation and model validation, as inaccuracies can result in clinically irrelevant or unsafe outcomes.
Third, many MVLM applications require addressing the challenge of imbalanced datasets, where a subset of particular conditions and patient demographics may be overrepresented. 
Effectively handling these challenges is vital for developing models that are not only accurate but also reliable and applicable in clinical settings.
\qinew{Lastly, unlike general vision-and-language models, which can tolerate some level of error without significant repercussions, MVLMs require a high degree of interpretability and trustworthiness since their outputs are used to support critical clinical decisions. This necessity for rigorous validation and performance underlines the importance of specialized methods tailored to the unique demands of the medical domain.}


Our aim here is to provide a dedicated survey that systematically reviews the current state of research on MVLMs. We provide a comprehensive overview of existing approaches, identify the key challenges, and suggest future research directions tailored to the medical context. Our goal is to promote interdisciplinary collaboration by bringing together AI researchers, clinicians, and healthcare professionals to develop innovative solutions that enhance clinical practice.
To address this need, we present a systematic survey of MVLMs across various medical multimodal tasks including:
\qinew{medical report generation (MRG)~\cite{messina2022survey,zhang2020radiology}, which aims to generate medical reports for radiographs by accurately localizing features, extracting information, and generating precise text; medical visual question answering (VQA)~\cite{gong2022vqamix}, wherein the system is expected to provide accurate answers to questions about medical images; medical multimodal diagnosis and prognosis~\cite{song2022multicenter,chen2021computer}, and text-guided medical image segmentation (MIS)~\cite{lu2024multi,lee2023text}, whereby text data like symptom descriptions and medical history combined with various modalities of the image to determine disease severity, assess outcomes, or segment medical images; and medical image-text retrieval (ITR)~\cite{abacha20233d}, which focuses on developing systems that efficiently retrieve relevant images or text, supporting clinical applications like diagnosis and education.}}

\xienew{The key motivation for providing a comprehensive overview across multiple domains of medical application is to enable a holistic analysis and comparison of the technologies employed across the entire field. In contrast, previous surveys have been more narrowly focused on specific tasks or methods within medical vision-language learning. For example, surveys on MRG \cite{hartsock2024vision,wang2024survey,pang2023survey,messina2022survey,liu2023systematic,liao2023deep,kaur2022methods} primarily review deep learning-based methods for generating radiology reports using multimodal data, but do not explore applications beyond MRG. 
Similarly, surveys on medical VQA \cite{al2021visual,lin2023medical,wang2022knowledge} analyze models that answer questions about medical images but do not extend to other key tasks. Likewise, multimodal learning surveys \cite{warner2024multimodal,cui2023deep} focus on fusing image and non-image data for tasks like disease diagnosis and prognosis but are limited in covering broader vision-language tasks, \qinew{such as medical VQA, needing the ability of both understanding and generation}. 
In addition, surveys on specific methodologies provide an in-depth look at individual models or approaches but remain confined to their respective areas. For instance, \cite{zhao2023clip} offers a comprehensive review of the application of CLIP models in medical imaging, but primarily focuses on this particular method. \cite{survey_wang2022medical} delves into deep learning methods for medical image segmentation, offering insights into this specific task, but it does not cover recent language-based segmentation approaches.
Overall, while each of these surveys provides valuable insights into specific areas, they remain task-specific or methodology-specific and do not generalize to the wider scope of vision-language models in medical AI.}

The main contributions of this study are as follows:
1) Unlike existing surveys that focus on specific aspects or tasks within the medical vision-language domain, we provide a 
longitudinal
and up-to-date review covering a wide range of key tasks in the last 5 years. This wide breadth allows us to identify overarching trends and interconnections across these tasks, offering a more integrated perspective on the field.
2) For each task, we delve into its significance in healthcare, challenges it poses, existing methods, datasets used for evaluation, experimental findings, and future research directions. By providing deep insights and perspectives, this paper can help readers better understand the current state of research and future directions in this field.

We organize the rest of the survey as follows. Section \ref{challenges} discusses the challenges faced in exploring MVLMs. Subsequent sections give details of five popular tasks, \ie, medical report generation (Section \ref{mrg}), medical visual question answering (Section \ref{vqa}), medical multimodal diagnosis and prognosis (Section \ref{diagnosis}), medical image segmentation (Section \ref{segmentation}), and medical image-text retrieval (Section \ref{retrieval}). We summarize and conclude the existing works and the future outlooks in Section~\ref{conclusion}.

\section{Medical Report Generation}\label{mrg}

\subsection{Task Description}

\begin{figure}
    \centering
    \includegraphics[width=0.48\textwidth]{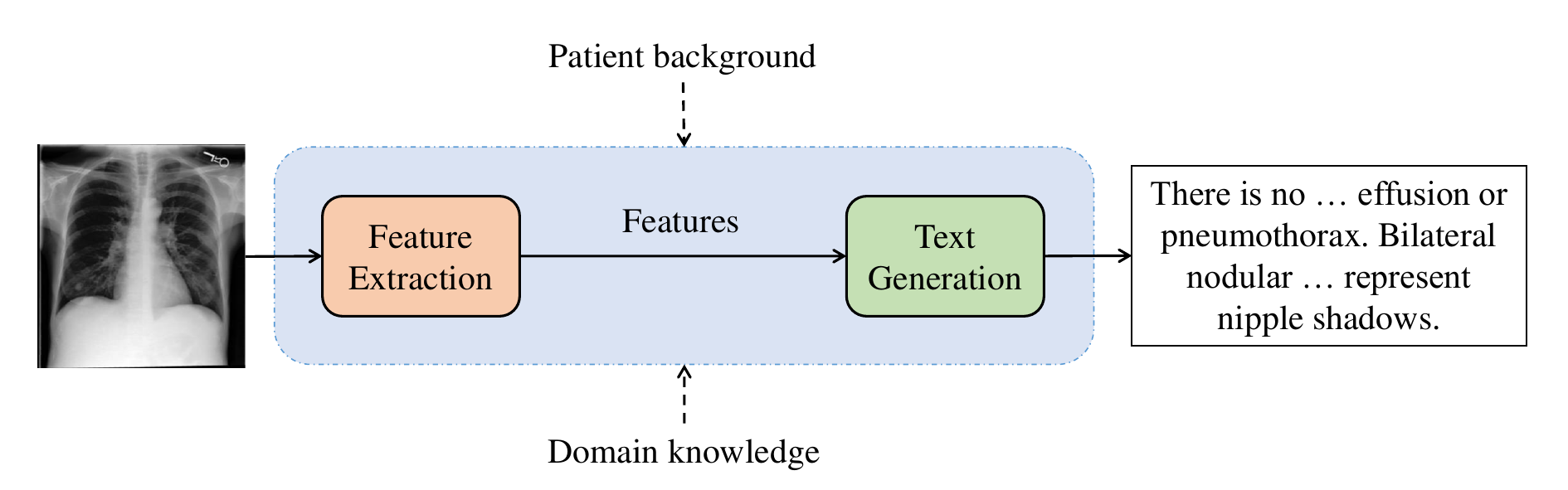}
    \caption{Typical framework of medical report generation.}
    \label{fig:mrg_framework}
\end{figure}

Medical report generation \cite{liu2022competence,delbrouck2022improving} aims to create detailed descriptions based on medical images, a critical and complex task in healthcare. Accurate interpretation of images is crucial for diagnosis and treatment, requiring significant time and attention from medical professionals. Automating this process with artificial intelligence reduces workload, improves efficiency, and helps address the shortage of specialized doctors. Current methods, as shown in Figure~\ref{fig:mrg_framework}, primarily use image captioning technologies to extract and decode image features for report generation.

\subsection{Visual Features from Images}

Medical images reveal the patient's pathological change,
providing critical information about the type, location, extent, and severity of the disease. This information forms an essential basis for clinical diagnosis, aiding doctors in making accurate diagnoses.
\qinew{Given a radiology image $X$, its a series of visual features $\mathcal{V}$ are extracted with an image encoder $f_v(\cdot)$, such as VGG \cite{simonyan2014very}, ResNet \cite{he2016deep} or Transformer~\cite{vaswani2017attention}. 
Mathematically, the process can be formulated as:
\begin{equation}
\mathcal{V} = \{\mathbf{v}_1, \mathbf{v}_2,...,\mathbf{v}_N\} = f_v(X), 
\end{equation} 
where $\mathbf{v}_i$ is the $i$-th patch embedding/token and $N$ refers to its total number.
}

\subsubsection{Feature Extraction}


\xienew{A number of neural networks have been commonly used in medical image analysis to extract key visual features. For example, DCNet~\cite{singh2022efficient} utilizes VGG16~\cite{simonyan2014very}, ResNet152V2~\cite{he2016deep}, and DenseNet201~\cite{huang2017densely}, which employ hierarchical feature extraction layers to focus on crucial visual patterns. MedSkip~\cite{pahwa2021medskip} integrates a modified HRNet with skip connections and attention modules, allowing the network to selectively prioritize important features during text generation while bypassing unnecessary information. Similarly, SGT~\cite{lin2022sgt} constructs a heterogeneous scene graph to capture key interactions between instruments and tissues in surgical scenes, enabling efficient local information processing and minimizing redundant data in the representation.}

\subsubsection{Feature Enhancement}

Visual features can be categorized into global and local features. Global features describe the entire image while local features focus on specific regions. When writing clinical reports, doctors consider both global and local features for a comprehensive judgment. An ideal MRG model should mimic this process, finding an organic way to integrate both types of features. However, the predominance of non-pathological regions in images often leads to bias, where models only focusing on global features might overlook local abnormalities. Thus, accurately capturing and describing abnormalities is crucial for effective MRG.

Recent research has introduced various methods emphasizing specific image regions. The contrast attention (CA) model~\cite{ma2021contrastive} identifies abnormal regions by comparing the current image with a healthy counterpart, highlighting visual characteristics of abnormalities. Class-incremental domain adaptation~\cite{xu2021class} uses a multi-layer transformer-based model, enhancing ResNet18 to process input images. Multi-task designs~\cite{wang2021self,szeskin2023liver,zhou2023advancing,tanida2023interactive} have proven effective in enhancing feature representation. Some methods directly identify abnormal regions through tasks like classification~\cite{tanida2023interactive} or segmentation~\cite{tanida2023interactive}. Other methods, \eg, MRM~\cite{zhou2023advancing}, use tasks such as mask prediction during pre-training to enhance information extraction.

\subsection{Cross-Modal Alignment}

For cross-modal learning models in MRG, which involve both vision and text, aligning these modalities enhances the conversion of image features to textual reports. A common approach is using a shared feature space, where visual and text data are mapped into the same space. In this space, images and text can be directly compared and matched. For example, attention-based methods~\cite{chen2022cross,wang2022cross,endo2021retrieval,sun2022lesion,gasimova2020spatial} automatically learn the important correlations between images and text, while Generative Adversarial Networks (GANs \cite{goodfellow2014generative}) achieve alignment through a generator and discriminator. Metric learning-based methods~\cite{tanwani2022repsnet,ni2020learning} align image regions with textual descriptions of anomalies by contrastive learning or ranking losses.

\subsection{Generating Report}

\begin{figure*}
    \centering
    \includegraphics[width=1.0\textwidth]{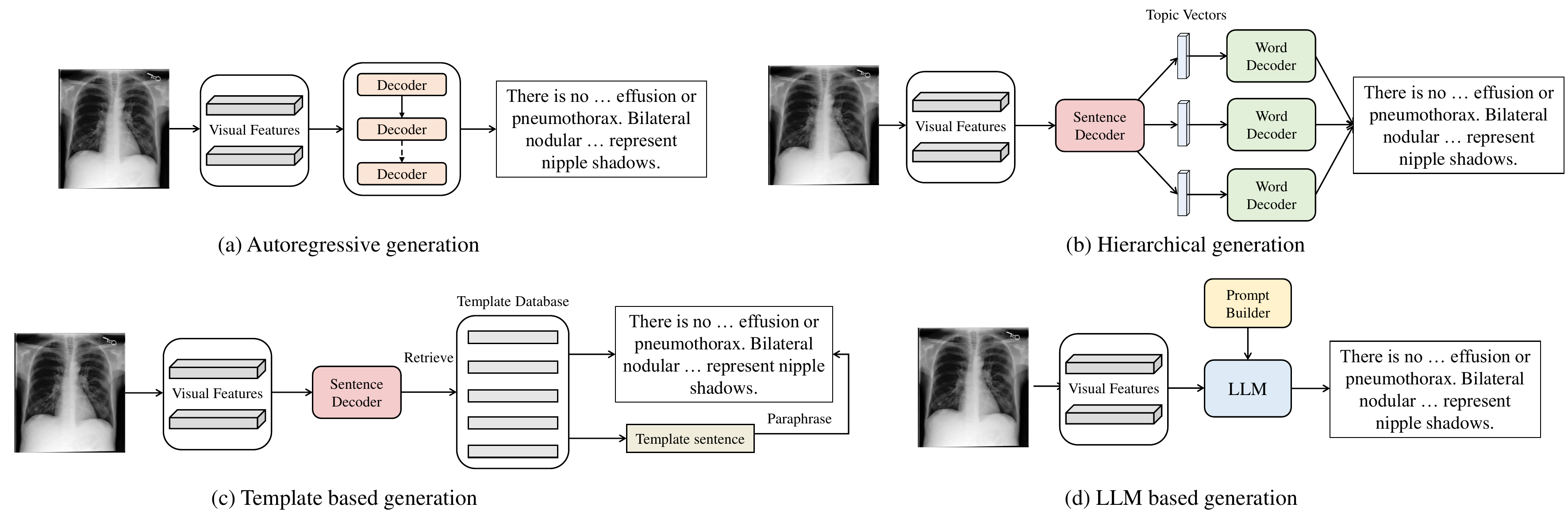}
    \caption{Four common ways of report generation: (a) Autoregressive architecture that generates report word by word based on LSTM, Transformer, etc; (b) Using hierarchical decoders that contain a sentence decoder and a word decoder; (c) Refer to sentences in the template database or fill the template sentences with specific information; (d) Generating reports relying on the comprehension and generation capabilities of large language models (LLMs).}
    \label{fig:text_generation}
\end{figure*}

\qinew{The decoder generates reports based on the probability given the visual features:
\begin{equation}
\hat{Y} = f_d(\mathcal{V}) = f_d(\mathbf{v}_{1},\mathbf{v}_2,\dots,\mathbf{v}_N),
\end{equation} 
where $\hat{Y}$ is the generated report and $f_d(\cdot)$ refers to the decoder.}
Three types of architectures for the decoder part are used to generate medical reports, as shown in Figure \ref{fig:text_generation}.

\subsubsection{Autoregressive Architecture}

The most straightforward decoding method is to use a recurrent neural network (RNN) such as LSTM \cite{huang2021deepopht,huang2022non} or GRU \cite{singh2022efficient} to generate the results word by word as shown in Figure \ref{fig:text_generation} (a). Some studies have incorporated attention mechanisms into this foundation \cite{najdenkoska2021variational}. 
%
Another type of autoregressive generation commonly used by existing methods is the Transformer architecture such as \cite{yan2023attributed,najdenkoska2022uncertainty,wang2022inclusive}. 
\xienew{Unlike LSTMs, Transformers \cite{vaswani2017attention} introduce a self-attention mechanism to address long-term memory loss, which refers to the difficulty in retaining information from earlier parts of a sequence when processing longer inputs. The self-attention mechanism allows Transformers to directly access and weigh all parts of the input sequence, regardless of distance, ensuring better retention of relevant information over long sequences.}
%
Works like \cite{wang2023metransformer} further modify the attention mechanism to fit the features received from the image encoder. Some of these works are designed for cross-modal enhancement or alignment~\cite{liu2021exploring,yang2022knowledge,yan2022prior,wang2023metransformer}. Others are designed to receive features extracted from auxiliary information \cite{yang2022knowledge,wang2023metransformer,zhang2023novel}. The auxiliary information here, whether generalized or specific medical knowledge, provides additional complementary descriptions of the original features. Such modified models allow for better attention to relevant visual or text features, processed from specific information or knowledge.

\subsubsection{Hierarchical Architecture}

In MRG, descriptions of normal and abnormal regions often vary greatly, necessitating detailed sentences to describe multiple image regions. To address this, many works use hierarchical decoders~\cite{zhou2021visual,zhang2020radiology,nooralahzadeh2021progressive,nguyen2021automated}. As shown in Figure \ref{fig:text_generation} (b), these include a sentence decoder to generate topic vectors and a word decoder to create sentences word by word based on these vectors. This approach ensures each sentence effectively conveys specific information, enhancing report interpretability and mitigating the limitations of a single decoder for long sentences. For instance, in~\cite{zhou2021visual}, the sentence decoder first generates a sentence, and subsequent sentences are created using previous sentence embeddings.

\subsubsection{Template-based Architecture}

The medical field uses standardized formats and criteria for reports to ensure their consistency and comparability. 
This standardization simplifies understanding and communication, allowing patterns in these reports to enhance and refine diagnostic descriptions and inspire new approaches to report generation.

To enable models to learn specific statements, a substantial amount of template data must be provided in advance. This data can come directly from original reports or be sentences that have been manually processed and checked. As shown in Figure \ref{fig:text_generation} (c), template-based report generation methods fall into two main categories: retrieval and hybrid retrieval-generation. In the retrieval-only approach, some studies select sentences directly from a database~\cite{kong2022transq} or modify them for accuracy~\cite{syeda2020chest,yang2021writing}. Other studies~\cite{pino2021clinically,han2021unifying} use database sentences as templates, filling in anomalous information to create the final report. For example, MedWriter~\cite{yang2021writing} employs a visual-language retrieval module to find the most relevant reports for an image and a language-language retrieval module to retrieve relevant sentences based on generated descriptions. The language decoder then combines image features with retrieved reports and sentence features to generate meaningful medical reports.

\subsubsection{LLM-based Architecture}
LLMs such as LLaMA \cite{touvron2023llama,touvron2023llama2}, GPT-series \cite{brown2020language,achiam2023gpt} and PaLM \cite{chowdhery2023palm}, trained on numerous texts, show powerful capability in natural language understanding and generation. They can generate high-quality textual reports based on user-provided structured or unstructured data, thus greatly reducing the time and effort of manual compilation. Traditional template generation methods have automated report writing to some extent, but are often overwhelmed when faced with complex and variable content. The emergence of LLMs provides a new solution in this area by automatically generating grammatical and logical text based on context and without the need for complex template design. As a result, LLMs are gradually being used for MRG~\cite{leonardi2023enhancing,wang2024llm,liu2024mrscore,wang2023r2gengpt}.

\qi{
The data fed into LLMs often needs to be preprocessed and formatted to textual prompts, ensuring that LLMs can understand as shown in Figure \ref{fig:text_generation} (d).
The critical aspect is to design the proper prompt to guide the LLMs generation process. For example, PromptMRG \cite{jin2024promptmrg} chooses to convert the diagnostic results of the classification branch into free-form reports by LLMs. \cite{ranjit2023retrieval} retrieves the relevant radiology text for a given image using multimodally-aligned embeddings and LLMs generate a report based on the retrieved text. Others like \cite{li2024prompt}, after identifying anatomical regions, generate sentences centered on key visual elements for structured reports. In this way, pre-trained LLMs can generate structured reports based on the anatomical region.
}

\subsection{Domain Knowledge}

\begin{figure}
    \centering
    \includegraphics[width=0.45\textwidth]{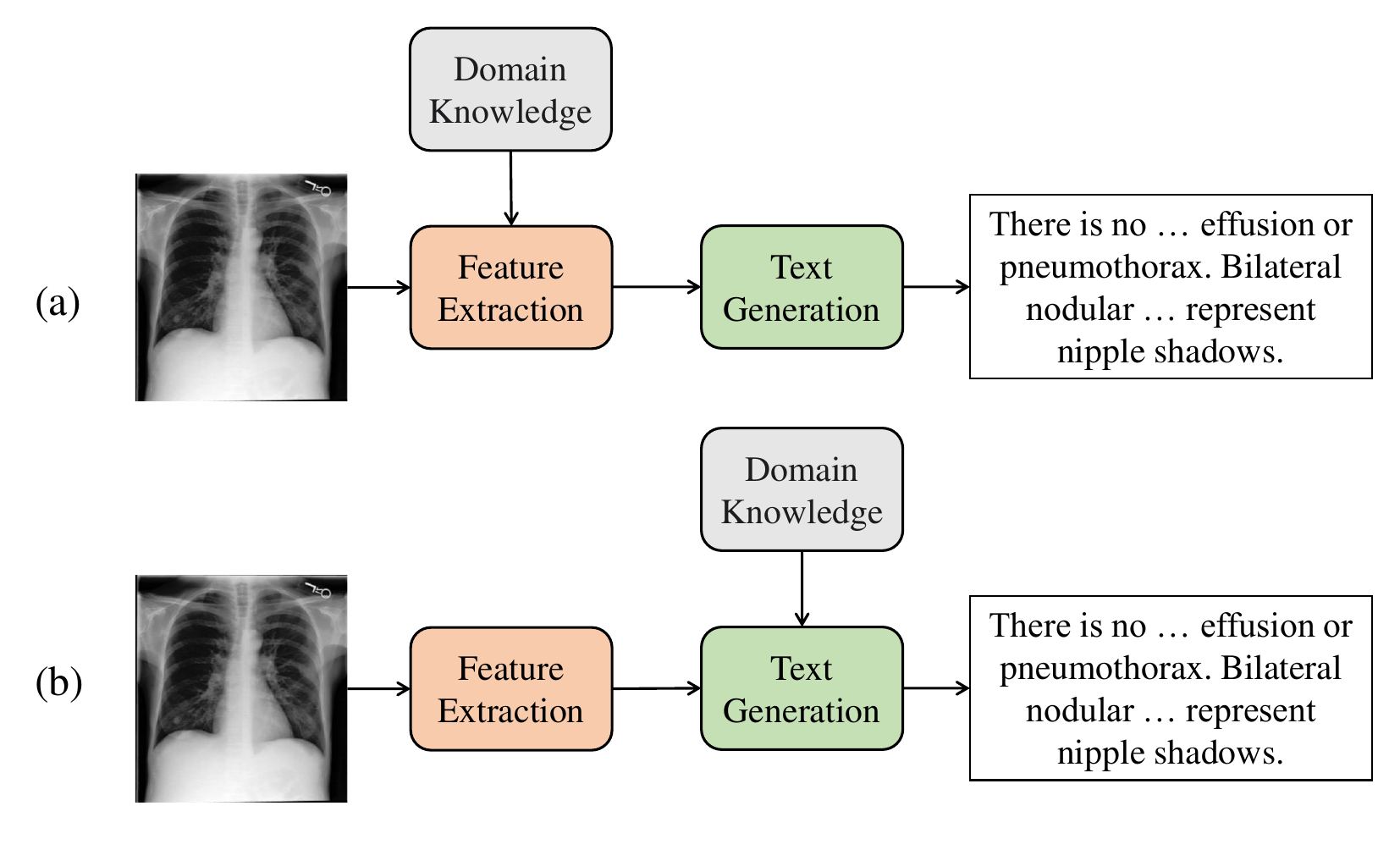}
    \caption{
    \qinew{Methods to leverage domain knowledge in medical report generation: (a) Domain knowledge assists in cross-modal feature fusion during feature extraction; (b) Domain knowledge supports the text generation process directly.}
    }
    \label{fig:mrg_knowledge}
\end{figure}

Despite significant progress in MRG in recent years, the availability of medical data remains limited. To overcome this, many studies have integrated additional domain knowledge, such as medical terms and knowledge graphs. Medical terms describe concepts, diseases, anatomical structures, and treatments, providing specific information in medical reports. Knowledge graphs are semantic networks that illustrate relationships between entities, with structural relationships representing different medical terms. These graphs can be pre-constructed based on medical knowledge or prediction-based using visual features. As shown in Figure~\ref{fig:mrg_knowledge}, models use domain knowledge in various ways, primarily in (a) cross-modal fusion and (b) text generation parts.

\subsubsection{Knowledge-based Cross-modal Fusion}

The introduction of medical terminology or knowledge graphs helps models selectively focus on specific visual features, facilitating cross-modal alignment. This alignment process can be categorized into two main approaches. Some studies~\cite{yang2022knowledge,yang2023radiology,li2023dynamic,yang2021joint,wang2022medical} encode image features and knowledge separately before fusion, then pass them to the decoder. Others~\cite{liu2021exploring,liu2021auto,cao2023mmtn,li2022self,huang2022non,li2022cross} encode them together, using attention mechanisms to filter important visual features or obtain knowledge representations for report generation. For instance, \cite{yang2022knowledge} uses a knowledge-enhanced multi-head attention mechanism to combine structural knowledge and visual features. \cite{liu2021exploring} employs a Posterior Knowledge Explorer to identify abnormal regions with disease keywords and a Prior Knowledge Explorer to leverage a disease knowledge graph. Besides, the AlignTransformer model~\cite{you2021aligntransformer} predicts disease labels from the input image without additional knowledge input.

\subsubsection{Knowledge-based Text Generation}

During report generation, domain knowledge aids models in making accurate judgments and conclusions. Medical terms help identify disease attributes, while knowledge graphs guide the description of diagnoses. Some models~\cite{zhu2023utilizing,dalla2023finding,zhang2020radiology} directly combine knowledge with visual features, \eg, \cite{zhu2023utilizing} uses attention mechanisms to integrate longitudinal and image embeddings during text generation. Others~\cite{huang2023kiut,zhang2023novel} process domain knowledge before generation. For instance, KiUT~\cite{huang2023kiut} uses an Injected Knowledge Distiller to extract valuable information from visual, contextual, and clinical knowledge. Some methods~\cite{zheng2023evidential,nishino2020reinforcement} predict keywords first and update them based on user feedback or expert corrections.

\subsection{Datasets and Results}

For public datasets, IU X-ray \cite{iu_xray_demner2016preparing} is the most commonly used and includes 7,470 frontal and lateral chest radiographs and 3,955 reports. The datasets MIMIC-CXR \cite{MIMIC-CXR_johnson2019mimic} and PadChest \cite{padchest_bustos2020padchest} contain 377,110 and 160,868 images, respectively, which can alleviate the problem of insufficient data quantity to some extent. In addition, some datasets have been used for image classification as a pre-training, intermediate or auxiliary task for report generation. Such classification datasets do not provide a report for each image but rather provide a set of clinical status or anomalies that are either present or absent in the images. An example is the CheXpert dataset \cite{chexpert_irvin2019chexpert}, which contains 224,316 images with a CheXpert classifier that annotates 14 labels in the natural language report as present, absent, or indeterminate. Due to the page limit, datasets, evaluation metrics, and experimental results are given in detail in supplementary~\ref{mrg_dataset}.

\subsection{Limitations and Insights}

As a promising research field, MRG still faces several significant shortcomings, including 1) insufficient data diversity and dataset size, 2) lack of reliable generation processes, 3) inadequate evaluation metrics and expert assessment, and 4) only English-centric generation.

Firstly, there is a lack of diversity and insufficient size in the dataset used for MRG studies. Most works focus on chest X-ray images and corresponding reports, leading to datasets with highly similar abnormal areas and fixed disease types and descriptions. This limits the models' ability to meet diverse clinical needs and reduces their generalization ability. Currently, other types of data and body parts are under-researched. Expanding research to include various modalities and body regions like~\cite{chen2023s4m} could lead to the development of more generalized and versatile models.

\qi{Secondly, it is non-trivial to ensure the semantic alignment among multimodal data. Models often learn from visual, textual, and various types of knowledge during training. However, it is hard to check whether domain knowledge, such as medical keywords, correctly matches the other modality data, such as images. LLMs offer strong language comprehension capabilities, which can provide a deep understanding of the semantic relationship between words, sentence structure, contextual dependency, and other information~\cite{wang2023r2gengpt,li2024prompt,jin2024promptmrg}. However, how to apply LLMs in the MRG systems remains an open question.
}

In terms of result checking, automatic evaluation typically uses traditional NLP metrics. However, NLP metrics like BLEU~\cite{papineni2002bleu} and CIDEr~\cite{vedantam2015cider} are not designed for the medical domain and cannot account for domain-specific phenomena. 
Moreover, due to the workload, it is challenging for MRG studies to perform expert assessments, yet these assessments are crucial for integrating models into clinical settings.
Thus, developing automatic and medical-specific evaluation metrics is a key direction for future research.

Finally, while English is predominantly used in vision-language datasets, it is not the native language for most of the global population. Research on monolingual data has led to limited performance and can create biases against non-English-speaking communities in medical applications. Narrowing down development in different languages is challenging due to lexical, grammatical, and cultural differences. Accurately representing medical terminology and clinical information is crucial. Although multi-language modeling often relies on machine translation~\cite{yang2024zeronlg}, it may suffer from semantic ambiguity and inaccurate terminology translation. Future research should focus on improving machine translation accuracy in the medical field and exploring ways to unify cross-language semantic representations directly, which can enhance the efficiency and quality of multi-language medical report generation like~\cite{wan2024med}.

\section{Medical VQA}\label{vqa}

\subsection{Task Descriptions}

\begin{figure}
    \centering
    \includegraphics[width=0.40\textwidth]{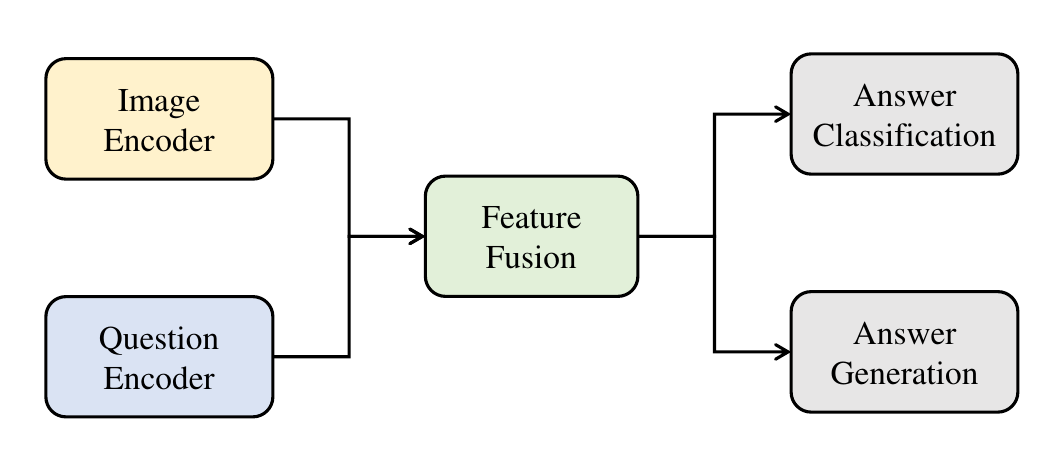}
    \caption{Framework of Medical VQA. The fusion module is used to process multimodal features. Using a classification or generation approach to get an answer depends on whether the question is open-ended or closed-ended.}
    \label{fig:vqa_framework}
\end{figure}

Medical visual question answering (VQA) aims to provide accurate answers to questions based on medical images. This process is challenging, requiring the analysis of both visual and textual information. Most architectures follow the framework used in general VQA.
\qi{
As shown in Figure \ref{fig:vqa_framework}, it typically involves an image encoder, a question encoder, a multimodal feature fusion method, and an answer prediction module.
\qinew{Concretely, given an image $X$ and a question $Q$, the model yields an answer $\hat{Y} = \{\hat{y}_1, \hat{y}_2,...,\hat{y}_{L}\}$, where $L$ is answer's length.}
QA settings can be open-ended or closed-ended. Closed-ended ones have limited answer choices and can be treated as classification tasks. Open-ended ones are more challenging, requiring a broader understanding and free-form responses.
}

\subsection{Enhanced Image Encoding}

While pretraining-finetuning paradigms improve vision-language model accuracy, the limited medical data and lack of annotations restrict pre-training on medical VQA datasets. Several studies~\cite{bao2021beit,wei2022masked,yu2022coca} have used pre-trained models on ImageNet for visual feature encoding, proving effective. Enhancing pre-training methods to handle data encoding efficiently is a key research focus in medical VQA. Current works mainly consider
contrastive learning \cite{liu2021contrastive,tanwani2022repsnet}, meta-learning \cite{cong2022anomaly,tascon2023localized}, and multi-task learning \cite{cong2022anomaly,bai2023revisiting}.

\subsubsection{Contrastive Learning}

Contrastive learning enhances the understanding of semantic relationships between images and text, improving the ability to answer questions accurately. MUMC~\cite{li2023masked} and M2I2~\cite{li2023self} use a pretraining-finetuning paradigm with a self-supervised framework, incorporating contrastive loss, masked language modeling, and image-text matching. They use image captioning datasets to learn unimodal and multimodal feature representations. Similarly, CPRD~\cite{liu2021contrastive} employs contrastive learning with unannotated images to create a general visual feature encoder, eliminating the need for auxiliary tasks and additional labeling. RepsNet~\cite{tanwani2022repsnet} is designed specifically to align images and text contrastively during encoding.

\subsubsection{Meta-learning}

Directly fitting models from the natural domain to the medical domain often leads to overfitting due to high data variance and scarcity. To address this, researchers employ meta-learning to use raw medical data better. VQAMix~\cite{gong2022vqamix} mitigates data limitations through a cross-modal MixUp of images and questions. MMQ~\cite{do2021multiple} generates meta-annotations for Model-Agnostic Meta-Learning (MAML) training. It produces meta-models with robust features for medical VQA tasks and handles noisy labels through auto-annotation without needing additional data.

\subsubsection{Multi-task Learning} 

Multi-task learning can improve the generalization ability of a model by jointly learning on multiple related tasks, which is indicated to be effective for VQA tasks~\cite{pollard2020visual,bai2023revisiting,cong2022anomaly}. For example, \cite{bai2023revisiting} is a multi-task model including answering and localization. Localization tasks that relate answers to instance-level localization can help learners deal with confusion when faced with various similar instruments and operations in surgical scenarios.

\subsection{Enhanced Question Encoding}

Question encoders convert text data into vector/embedding, capturing the semantic information. Widely used models for question encoders include LSTM, Bi-LSTM, GRU, and Transformers like BERT \cite{bert_devlin2018bert} and BioBERT \cite{lee2020biobert}. Transformers-based and BERT-based pre-training models, especially with self-supervised masked language modeling~\cite{mmBERT_khare2021mmbert,li2023masked,li2023self}, have proven effective in the medical domain. Skip-thought vectors~\cite{kiros2015skip} are also viable, learning semantic and contextual representations through dynamic vocabulary. For inputs, a hierarchical VQA framework~\cite{pellegrini2023rad} integrates current questions with historical questions and answers to ensure consistency. While image encoders have seen more innovation, developing text encoders for the medical domain enhances the understanding and accuracy of medical VQA systems.

\subsection{Fusion Methods}

Fusion methods combine visual and textual features extracted by encoders, modeling the semantic links between them. As a key part of VQA, an efficient fusion method uses the complementary strengths of multiple features for accurate question answering. Common fusion methods include the attention mechanism and multimodal pooling.

\subsubsection{Attention Mechanism}

The attention mechanism, widely used in various tasks like classification, detection, and segmentation, mimics human observation by focusing on vital local information and combining it to form an overall impression. It allows models to assign different weights to input parts, extracting critical information for accurate judgments, thereby improving target recognition and classification.

In VQA problems, Stacked Attention Network (SAN)~\cite{yang2016stacked} and Bilinear Attention Networks (BAN)~\cite{kim2018bilinear} are commonly used attention methods like models in \cite{liu2022medical,zhan2023debiasing}. SAN uses question features as queries to rank image regions related to the answer, which focuses on crucial regions through multiple attention layers to filter out noise and identify relevant areas. BAN extends co-attention to bilinear attention, using bilinear interaction graphs and low-rank bilinear pooling to represent question and image features jointly.
Multi-head attention mechanisms are often used in the pre-training phase, with Transformer-like model architectures incorporating attention mechanisms such as~\cite{mmBERT_khare2021mmbert,li2023self,li2023masked}. 
For instance, the MF$^2$-MVQA~\cite{song2023mf} uses a CNN encoder to obtain multi-stage feature maps from medical images, which are then added to each Transformer layer stage by stage. This method leverages multi-scale information by the attention mechanism, avoiding confusion of visual features at different stages.

\subsubsection{Multimodal Pooling}

Multimodal pooling is a key technique for fusing visual and textual features, typically using concatenation, summation, or element-wise product. Models like Q2ATransformer~\cite{liu2023q2atransformer} and hi-VQA~\cite{pellegrini2023rad} integrate image and question features through concatenation, a simple method that effectively reduces information loss. However, these basic operations may not capture the complex relationships between modalities, and high-dimensional input vectors can make the dot product computationally expensive.

Some studies propose more efficient bilinear pooling methods to provide enriched multimodal representations of visual and textual features. Multimodal Compact Bilinear~(MCB) pooling~\cite{fukui2016multimodal} projects features into high-dimensional vectors and performs multiplicative convolution in Fourier space. However, MCB's high-dimensional nature can limit its applicability. To address this, the multimodal Low-rank Bilinear (MLB) algorithm~\cite{kim2016hadamard} reduces the rank of bilinear pooling using the Hadamard product. QC-MLB~\cite{vu2020question} uses the MLB module to query an image with a written question, which emphasizes question features with a multi-glimpse attention mechanism. Besides, MedFuseNet~\cite{sharma2021medfusenet} employs two attention modules to allow features from different modalities to interact twice, which combines the attended image and question features using Multimodal Factorized Bilinear~(MFB) pooling~\cite{yu2017multi}.

\subsection{Closed-ended vs. Open-ended}

\begin{figure}
    \centering
    \includegraphics[width=0.40\textwidth]{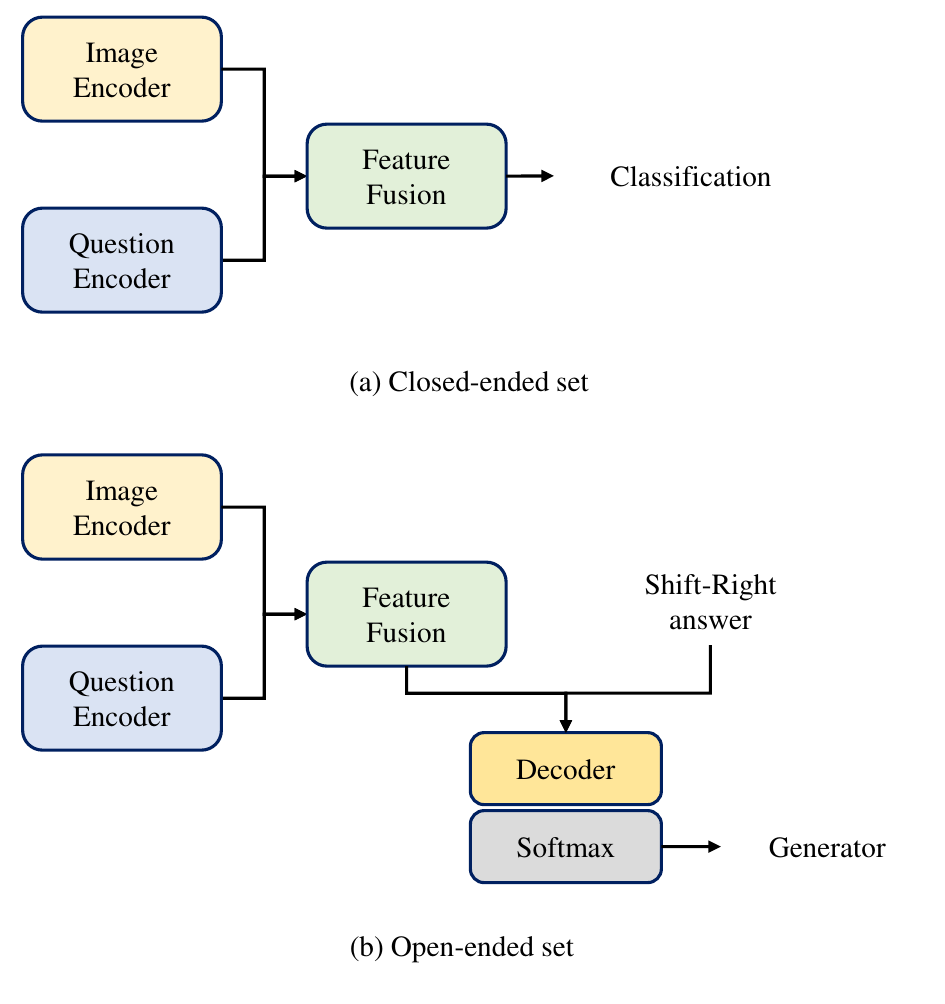}
    \caption{
    \qinew{Methods of medical VQA: (a) The classification architecture for closed-ended questions uses feature fusion from image and question encoders to classify the answer; (b) The generation architecture for open-ended questions employs feature fusion, followed by a decoder and softmax layer, to generate a more detailed answer.}
    }
    \label{question_answer}
\end{figure}

Among the reviewed papers, most models use classifiers to output answers, while a few treat question answering as a text generation task (see difference in Figure~\ref{question_answer}). The effectiveness of these methods depends on the length of the ground-truth answers. Classification is simpler and advantageous within a smaller answer space, but it struggles with longer, open-ended questions or complex answers like detailed lesion descriptions. Generative methods, meanwhile, often produce non-existent answers, leading to lower accuracy.

Some methods~\cite{ren2020cgmvqa,sharma2021medfusenet} use a switching strategy, combining classification for closed-ended questions and generation for open-ended ones. This approach requires sophisticated models but provides accurate results for different question types. For example, \cite{liu2023q2atransformer} unifies handling both question types by integrating classification and generation. It uses learnable candidate answer embeddings during decoding to query each answer class and interact with fused features by attention mechanisms. This process effectively reduces the search space for answers. It also has the benefits of the generation-based open-ended framework.

Several studies propose innovative approaches beyond combining classification and generation to accurately generate answers. \cite{van2023open} uses a mapping network to convert image features into learnable tokens, which serve as visual prefixes fed to the language model along with question features. These visual prefixes guide the language model to generate accurate answers by effectively communicating medical image information. \cite{liu2022medical} integrates a Question-Conditioned Reasoning (QCR) module to capture question attention information, which enhances reasoning skills, and a Type-Conditioned Reasoning (TCR) module to learn the difference between question types.

\subsection{Datasets and Results}

The most widely used datasets for medical visual language quizzing are VQA-Med \cite{hasan2018overview,abacha2019vqa,ben2021overview}, VQA-RAD \cite{lau2018dataset}, PathVQA \cite{he2020pathvqa}, and SLAKE \cite{liu2021slake}. Each dataset contains pairs grouped into certain categories. The datasets, evaluation metrics, and experimental results are given in detail in supplementary \ref{vqa_dataset}.

\subsection{Limitations and Insights}

Current medical VQA models still face several limitations, including insufficient high-quality data, lack of support for multi-round dialogue, ineffective text encoding for question analysis, limited generalization ability across datasets, and need for improved interpretability and credibility.

First, low-quality data is a major limiting factor.
We find that current medical VQA models fall short of being able to assist the healthcare system effectively. The main issue is the low quality and limited scope of data, which often includes impractical questions, \eg, ``Does the image contain the left lung?'', 
that are not typically asked by doctors or patients. Instead, they care about disease presence and progression. Such data with low clinical relevance does not help the model in making accurate disease judgments.

The structure of medical VQA models is largely standardized, with a significant focus on innovating image encoders. In contrast, fewer studies propose new methods for encoding textual features or extracting question information. The limited variety of questions current models can handle suggests that text encoding in medical VQA may not be as optimal as it could be.

Besides, the generalization ability of medical VQA models needs improvement. Many models are tailored to specific datasets and cannot be easily applied to other medical datasets, resulting in suboptimal accuracy when faced with unknown situations. The advent of large language models offers the potential for enhancing question encoders. With advancements in NLP and medical knowledge graphs~\cite{wang2022knowledge}, future models can achieve a deeper understanding of the medical text and improve question analysis capabilities.

Medical VQA models lack support for multi-round dialogue, a critical aspect of real-world medical interactions where patients and doctors engage in dynamic, ongoing conversations. Currently, these models handle only single questions and responses, without tracking context or dialogue coherence. This limitation restricts their practical application, as effective medical consultations often require multiple rounds of dialogue to fully understand symptoms or clarify diagnoses. Even in the natural domain~\cite{pang2023multi}, visual dialogue for multi-round interactions is rarely explored.

Finally, advancing the interpretability and credibility of medical VQA systems is crucial for future research. Ensuring that the model's answers are understandable and accurately localized will build trust among medical experts and encourage clinical adoption. Some studies \cite{bai2023cat,tascon2023localized} incorporate localization and labeling of pathological regions while answering questions, using a multi-task strategy to verify answer accuracy, which is promising but less explored in medical VQA.

\section{Medical Multimodal Diagnosis and Prognosis}\label{diagnosis}

\subsection{Task Description}

Medical multimodal diagnosis identifies the type, cause, severity, and treatment options for a disease by analyzing clinical manifestations, test results, and medical images, requiring a comprehensive judgment based on symptoms. Medical multimodal prognosis predicts disease progression, treatment effectiveness, risk of complications, and patient outcomes.

Compared to unimodal diagnosis and prognosis, medical multimodal diagnosis and prognosis (integrating reports, labels, and other textual information) offer more comprehensive insights, leading to more accurate diagnostics. With data from various modalities, a better understanding of a patient's health status can achieved.

\subsection{Methods}
\begin{figure}
    \centering
    \includegraphics[width=0.5\textwidth]{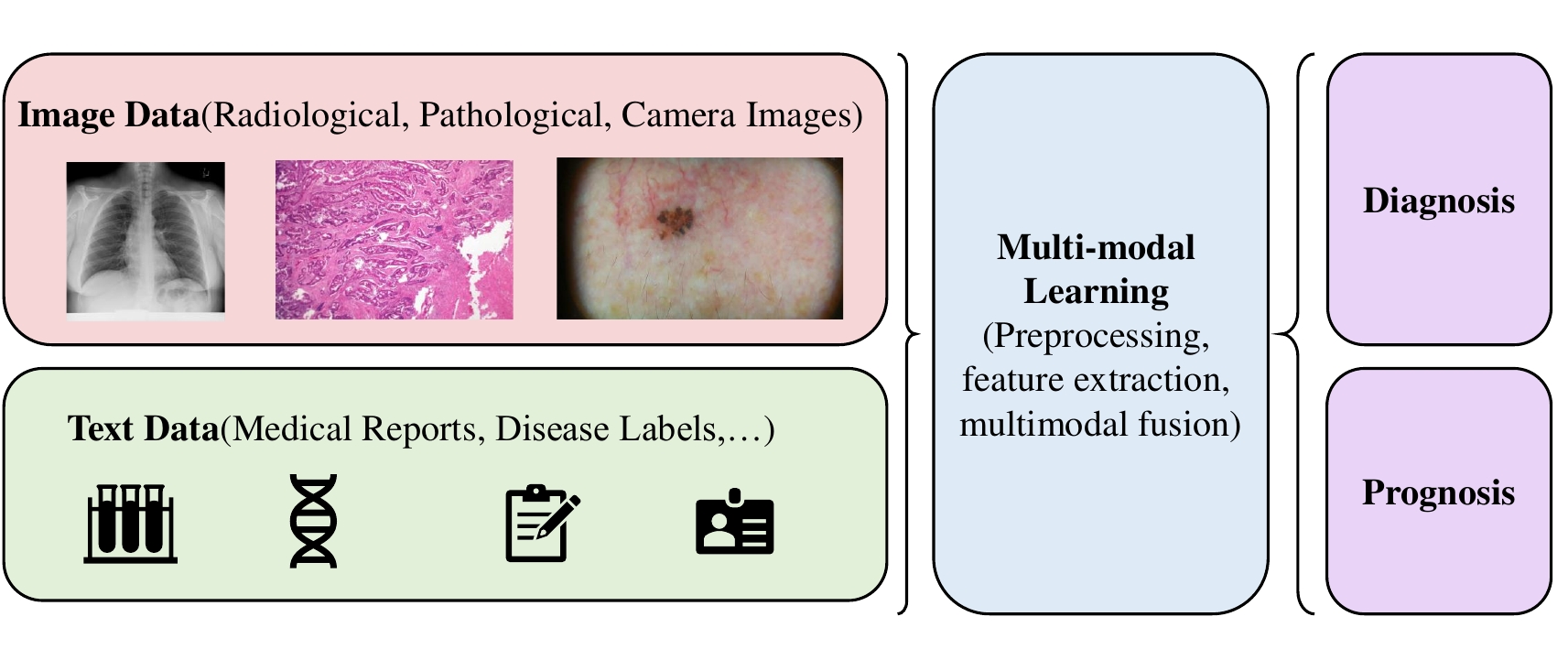}
    \caption{Framework of medical multimodal diagnosis and prognosis.}
    \label{diagnosis_framework}
\end{figure}

Many studies have succeeded in medical diagnosis or prognosis using unimodality \cite{deng2020deep,solares2020deep}. However, effective fusion of multimodal data is not an easy task, as different clinical modalities contain different information and have different data formats. We show the data types involved and the process of medical multimodal diagnosis and prognosis in Figure \ref{diagnosis_framework}.

Commonly used images have different dimensionality, \eg, 2D and 3D. Text data usually include medical reports, disease labels, and diagnosis records. Images are larger and denser, while text data are sparser and lower dimensional. For example, 2D pathology images provide the microscopic morphology of a tumor, whereas 3D CT or MRI radiology images provide macroscopic and spatial information \cite{cui2023deep}. Therefore, these heterogeneous data formats require models to use different processing and feature extraction methods. 

\subsubsection{Multimodal Fusion}

In medical multimodal diagnosis and prognosis, textual information provides detailed descriptions of aspects such as medical records, symptoms, and treatment plans. In contrast, images display intuitive visual information like lesions and anatomical structures. 
Currently, the commonly used fusion method relies on the cross-attention mechanism.

The main categories of short text that are often introduced in medical multimodal diagnosis and prognosis models are text prompts and disease labels \cite{gundel2021robust,mehta2022end}. For example, \cite{mehta2022end} uses a bag-of-words approach to transfer the input image into words or patches. Self-attention is then used to encode inter-word and inter-bag relationships in a hierarchical way.

Long textual information commonly used are medical reports and diagnostic records \cite{van2020towards,qiu2021modal,monajatipoor2022berthop,ichinose2023visual,zhong2023ariadne}. These models strategically leverage the contextual relationships inherent in these modalities. For instance, \cite{van2020towards} establishes a unified feature representation by integrating prior patient information from Electronic Health Records (EHR) with associated X-ray scans.
Some other studies~\cite{lu2022m,liu2023multi} also explore the intrinsic relationships between multiple medical tasks based on the fusion of multimodal interaction information. They make full use of complementary information between tasks and combine the power of multiple modalities and multiple tasks to improve model performance.

\subsubsection{Image-Text Contrastive Learning}

Beyond fusing information from different modalities, a significant portion of research focuses on comparing images and texts, often by using contrastive learning \cite{chauhan2020joint,bhalodia2021improving,kim2022weakly,lei2023clip,chen2023gsdg}. For example, \cite{chauhan2020joint} uses the joint embedding loss to associate images with texts. \cite{bhalodia2021improving} introduces contrastive functions to align attribute features with region features along with the loss function for attribute classification of region features.

Some studies enhance image-text contrastive learning using models like CLIP and ChatGPT \cite{liu2023chatgpt,lei2023clip,pellegrini2023xplainer}. For instance, \cite{liu2023chatgpt} introduces a CLIP-based zero-shot medical image classification framework with ChatGPT to enhance contrastive learning. The model queries LLMs for additional cues and knowledge, \eg, disease symptoms, that improve the alignment between medical images and textual descriptions. In addition, prompts like ``Q: According to published literature, what are useful medical visual features for distinguishing \{Diagnostic Category\} in a photo?'' 
\xienew{help the model to refine its focus on medically relevant visual features.}

\subsubsection{Diagnosis and Prognosis Strategies}

The exact method of medical multimodal diagnosis and prognosis is determined by the type of disease \cite{wu2020automatic}. 
For instance, \cite{chen2021computer} introduces a multi-view ensemble learning approach incorporating a voting mechanism to enhance the efficacy of the model. Specifically, it integrates three diagnostic results derived from a 3-view dataset, comprising thyroid nodule ultrasound images, medical features extracted from U-Net \cite{ronneberger2015u} and relevant features selected by Max-Relevance and Min-Redundancy (mRMR).

Some studies also support medical multimodal diagnosis and prognosis with the help of graphs. \cite{sekuboyina2021relational} constructs a multimodal knowledge graph based on the chest X-ray images and their labels. Then, the multi-label classification is reformulated as a link prediction problem within this knowledge graph. \cite{song2022multicenter} proposes a multi-center attention graph. Each node in the graph represents a subject. This graph provides information on diverse data sources, disease states of training samples, gender, and equipment type information, facilitating the exploration of the influence of these factors on the graph convolutional network (GCN). 

\begin{figure*}[t]
    \centering
    \includegraphics[width=0.78\textwidth]{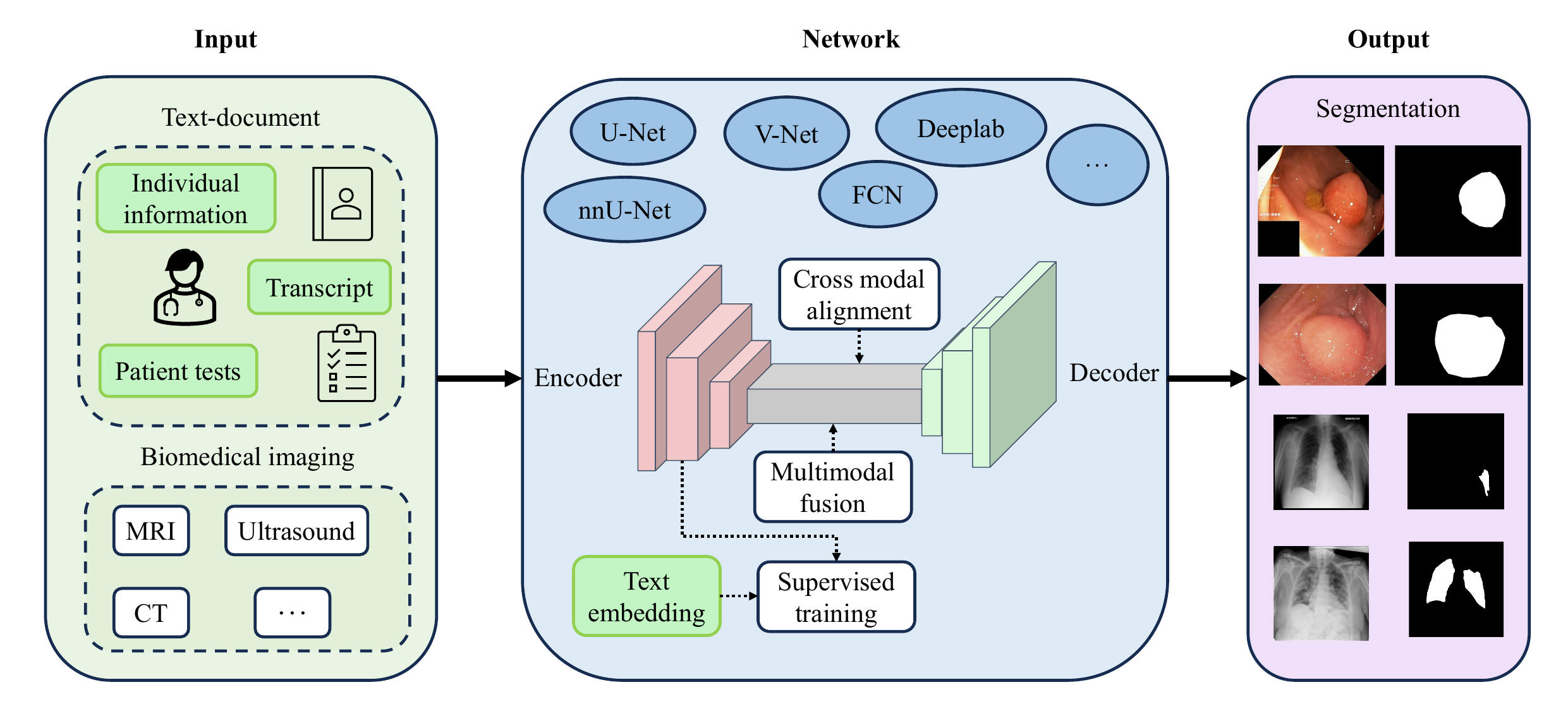}
    \caption{Framework of Text-guided Medical Image Segmentation. Segmentation networks usually use a fully convolutional network with an encoder-decoder structure. The encoder is used to extract features. Text features are combined with image features through fusion, alignment, or training strategies. The decoder is used to recover the extracted features to the original image size and output the segmentation result, typically U-Net \cite{ronneberger2015u}, FCN \cite{long2015fully}, Deeplab \cite{chen2014semantic}, \etc~Patient tests are a series of diagnostic procedures and examinations conducted to assess the health status, diagnose conditions, and monitor the progress of patients. These can include blood tests, imaging studies, biopsies, and other laboratory or clinical evaluations.}
    \label{segmentation_framework}
\end{figure*}

\subsection{\qi{Limitations and Insights}}

Researchers seek to improve diagnosis and prediction by integrating images and text through better feature fusion methods, but current models still struggle with these. First, it is non-trivial to verify whether the chosen feature fusion strategy is suitable for the current medical task. Currently, common feature fusion methods include concatenation, weighted summation, attention mechanism, \etc, whereas it is not easy to determine which method is more suitable for a specific task and dataset. Second, imbalances and mismatches between image and text data can cause some feature fusion methods to overly rely on one modality, neglecting others~\cite{zhong2023ariadne}, and creating redundant features that reduce representation ability. Besides, medical multimodal data is often scarce, especially for rare diseases~\cite{liu2023chatgpt}, and there's an imbalance between different data types, making it hard for models to learn and generalize, leading to inaccuracies in certain cases~\cite{chen2021computer}.


In future medical multimodal diagnosis and prognosis, the focus of research continues to revolve around how to fuse medical data from different modalities more effectively. Firstly, new techniques can be proposed in terms of feature extraction, attention mechanisms and cross-modal information transfer \cite{gundel2021robust,mehta2022end}. Secondly, effective and efficient usage of limited data is another practical and necessary issue. To reduce the limitations imposed by data scarcity, data enhancement methods can continue to be investigated to fully utilize available data while mining more data resources. Finally, explainability is another pressing issue in medical multimodal diagnosis and prognosis. Lack of transparency has been identified as one of the main barriers to deploying deep learning approaches in clinical practice \cite{panagoulias2024evaluating}.

\section{Text-guided Medical Image Segmentation}\label{segmentation}

\subsection{Task Description}

Medical image segmentation \cite{aleem2024test} aims to segment structures or tissues in medical images. The image segmentation task is mainly divided into semantic segmentation and instance segmentation. Semantic segmentation is the pixel classification of an image. 
\qinew{Given an input image $X$ with $K$ pixels, the objective of image segmentation is to assign labels $Y_i$ to every pixel $X_i$.}
Instance segmentation is required to identify different object instances that belong to the same category, further developing semantic segmentation. 
%
Segmenting fine structures of images enables quantitative analysis of tissue volumes~\cite{huang2023tissue}, aids clinicians in diagnosis and prognosis~\cite{zhou2024self,sonia2024breast}, and treatment planning~\cite{zhou2024deep,wang2023root}. 


\xie{With the increasing application of vision-language models and LLMs in the medical field, medical image segmentation models have started to introduce textual information, which is used for multi-modal fusion, cross-modal alignment, or supervised training. \qi{The process of such a text-guided medical image segmentation is in Figure~\ref{segmentation_framework}.}}

\subsection{Methods}

\begin{figure}
    \centering
    \includegraphics[width=0.5\textwidth]{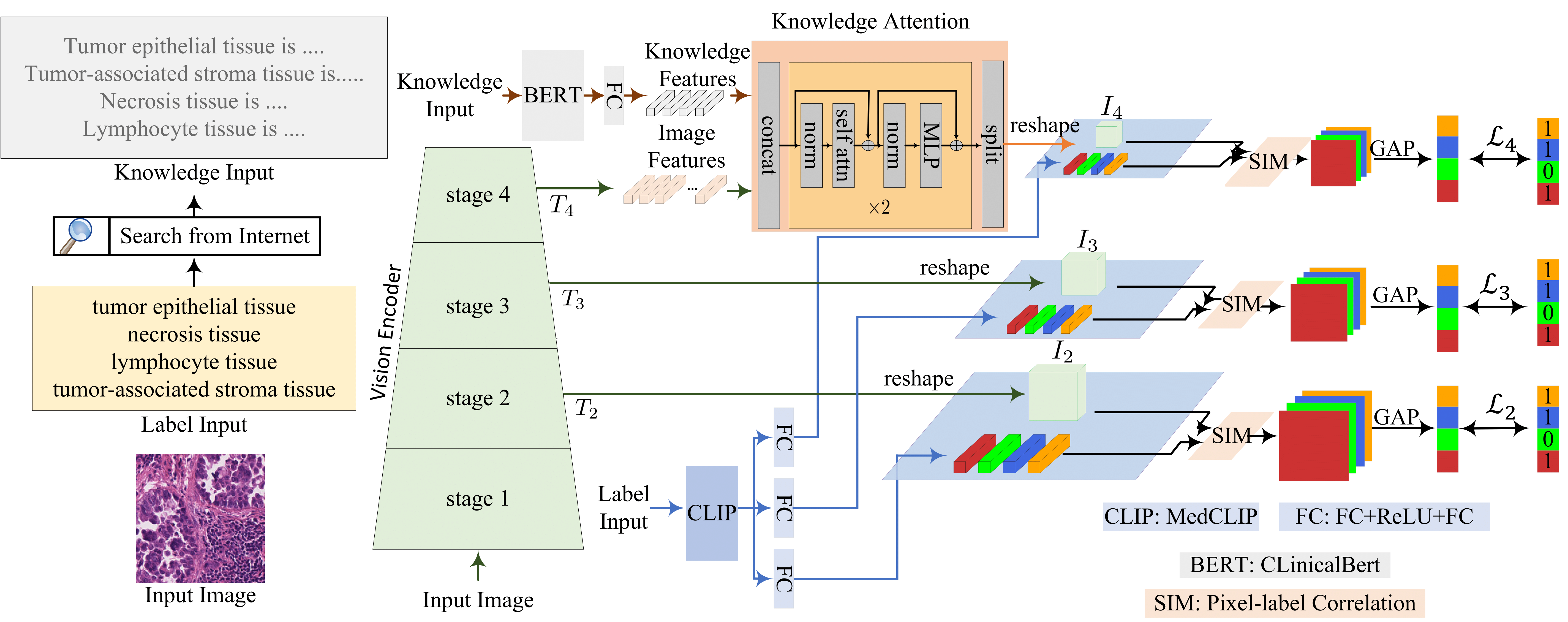}
    \caption{Framework of TPRO \cite{zhang2023tpro}. It integrates BERT-embedded textual descriptions of tissue subtypes with image features via a knowledge attention module while aligning CLIP-produced semantic representations and refined image features to produce segmentation maps.}
    \label{TPRO}
\end{figure}

Current segmentation methods either leverage cross-modal affinity or deeply fuse the language modality for segmentation. 
First, the former approaches \cite{zhang2023tpro,aleem2024test} capitalize on the semantic compatibility between pixel features and the textual descriptions of target classes to extract pixel-wise labels. 
\xie{Notably, TPRO \cite{zhang2023tpro} uses BERT to embed textual descriptions of tissue subtypes as knowledge features, which are directly integrated with image features through a knowledge attention module. Additionally, it utilizes similarity maps between refined pixel-wise image features and textual representations of target classes produced by CLIP to generate segmentation maps.} Figure \ref{TPRO} illustrates TRPO's architecture. Recent methods like \cite{aleem2024test} further process the image-text similarity map using SAM \cite{kirillov2023segment} to refine the segmentation outputs.

\begin{figure}
    \centering
    \includegraphics[width=0.5\textwidth]{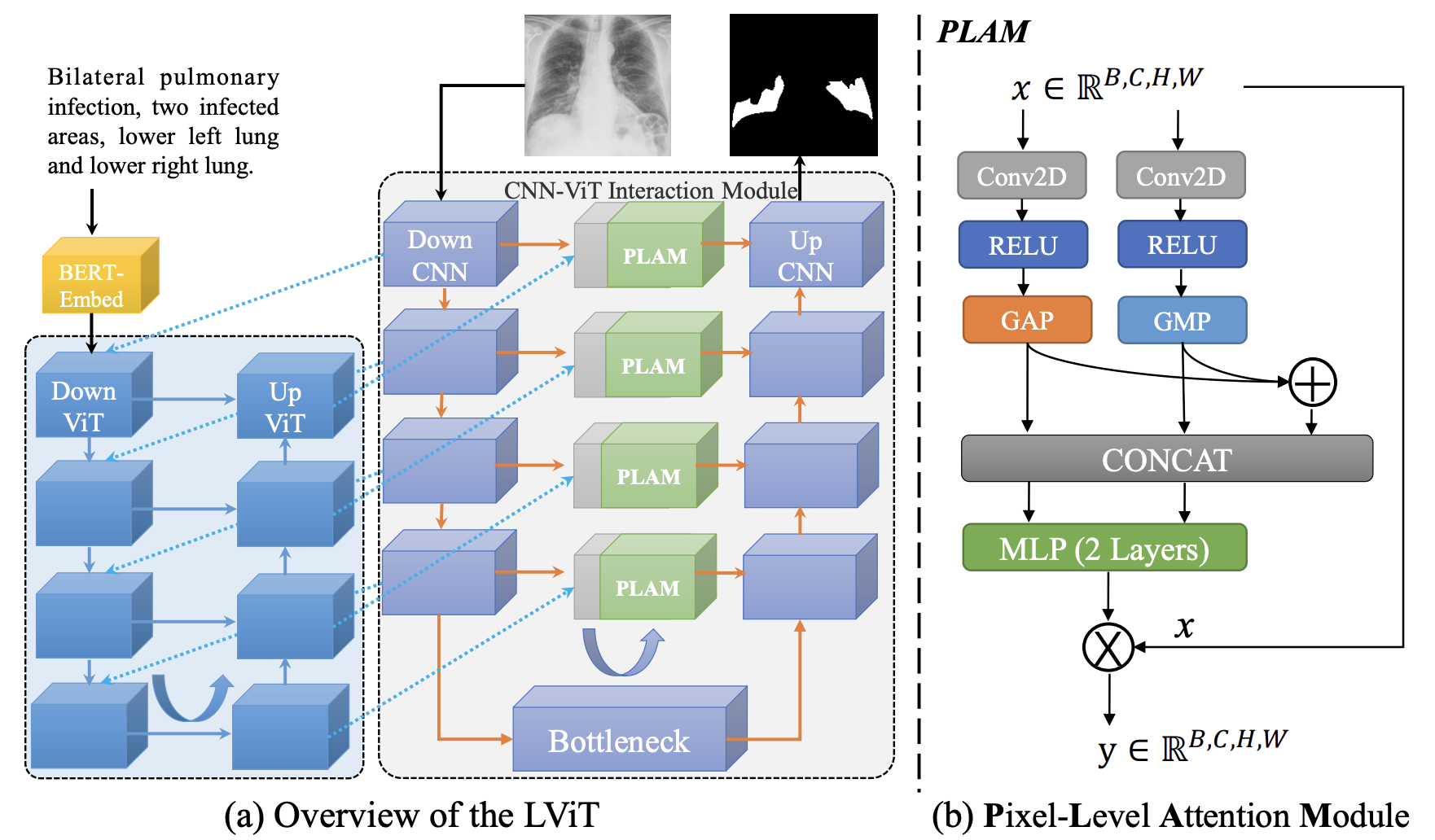}
    \caption{LViTin \cite{li2023lvit} combines a U-shape CNN branch with a U-shaped ViT branch.}
    \label{LViT}
\end{figure}

Second, studies (\eg, \cite{liu2023clip,shen2024segicl}) fuse the image and text features to enhance local representation for semantic segmentation. CLIP-Universal Segmentation \cite{liu2023clip} fuses text and image semantics via a Multilayer Perceptron (MLP) to enable universal segmentation. Similarly, SegICL \cite{shen2024segicl} incorporates texts in the in-context learning framework for universal segmentation. LViT \cite{li2023lvit} adopts U-shape fusion between two modalities, as shown in Figure \ref{LViT}. \cite{han2023multiscale} designs a prior prompt encoder (PPE) incorporating text prior prompts to generate multimodal features. Then, the Multiscale Feature Fusion module (MSFF) combines the features of PPE to generate multiscale multimodal features. Finally, the UpAttention module refines the prediction results by fusing image and textual features. Transforming single-scale features into multi-scale representations not only effectively solves the semantic gap between natural and medical data, but also improves the accuracy of prediction masks.

Third, another paradigm proposes to use textual information to formulate supervisory training objectives \cite{segre2022shape}. A recent method \cite{kunhimon2024language} introduces text-guided contrastive learning to enhance the semantics of visual features. RecLMIS \cite{huang2024cross} adopts conditioning masked reconstruction to predict masked texts from images and vice versa.

\subsection{Limitations and Insights}

Current text-guided medical image segmentation models still struggle with some limitations. First, most segmentation models are based on general class-level texts, which have not yet taken into account the differences among patients. Future studies could consider further refinement of the text extracted from patient medical records.
Second, the image features of different modalities vary greatly. For example, there are significant differences in visual features and pixel distributions between CT and MRI images. The same medical entity may be described differently in different modality images. Currently, the available text for these images tends to be the same. More specific text descriptions corresponding to different types of images can be further explored.
Third, segmentation efficiency needs to be further improved. Due to the introduction of LLMs and diffusion models \cite{zhao2024dtan,dong2024diffusion}, the segmentation speed is relatively slow. Thus, future research can improve the speed of segmentation models through parameter pruning and quantization, or efficient LLMs.

\section{Medical Image-Text Retrieval}\label{retrieval}
\subsection{Task descriptions}

\xie{Diagnosing complex cases or performing differential diagnoses, where similar visual cues must be distinguished, presents challenges for clinicians and often requires consulting various resources, which is time-consuming and disrupts workflow \cite{mbilinyi2021retrieving}. To address this, recent advances in automatic medical cross-modal retrieval methods aim to simplify the search process in large-scale, complex databases.

Traditional medical retrieval methods primarily focus on image-only retrieval but lack integration with textual information, resulting in suboptimal outcomes, especially in multi-label medical data where text is essential for accurate categorization. To enhance retrieval precision, recent methods \cite{mbilinyi2021retrieving, SECMR_ding2023semantic, MMDL_yu2021multimodal, DMCAH_zhang2022deep, MultiModal_Medical_zhang2023multi, text-guided_serieys2022text} incorporate descriptive text, enabling more precise image retrieval and flexible cross-modal retrieval by learning shared representations for both images and text, thereby bringing semantically similar instances closer together.
} 

\subsection{Methods}
\begin{figure*}
    \centering
    \includegraphics[width=\textwidth]{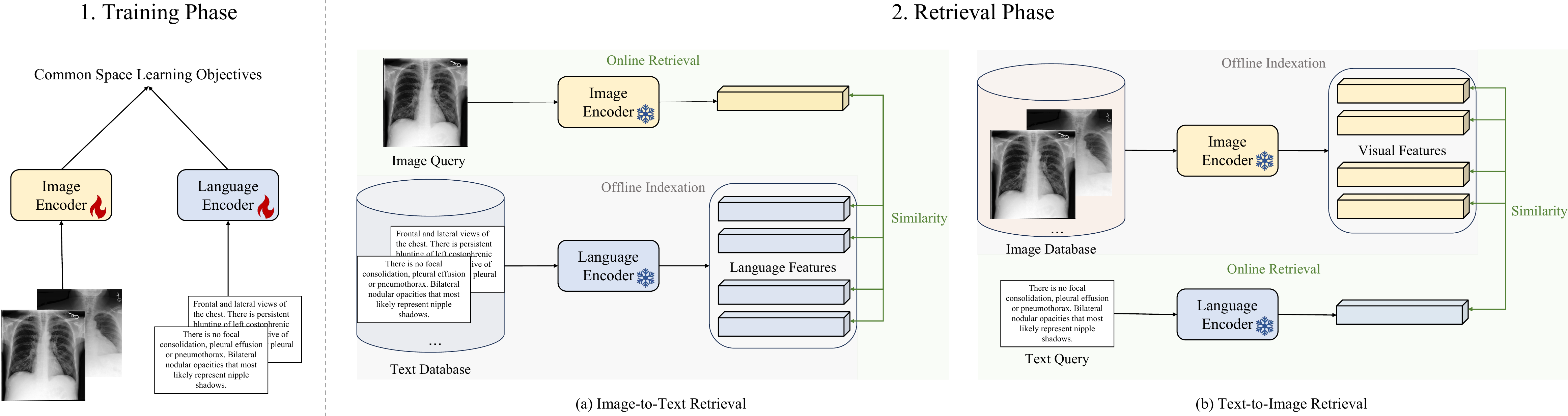}
    \caption{Framework of Medical Vision-Language Retrieval.}
    \label{fig:retrieval}
\end{figure*}

\begin{table*}[!t]\label{tab:retrieval}
\centering
\caption{Overview of current medical image-text retrieval methods.}
\renewcommand\arraystretch{1.25}
\resizebox{0.95\textwidth}{!}{
\begin{tabular}{c|c|cc}
\hline
\textbf{Feature Representation} & \textbf{Method}                 & \textbf{Highlights}                                                                                                                              & \textbf{Dataset}        \\ \hline
\multirow{3}{*}{Binary}         & ~\cite{category_zhang2022category}                        & Category Supervised Hash Learning                                                                                                                & MIMIC-CXR~\cite{MIMIC-CXR_johnson2019mimic}               \\ \cmidrule{2-4} 
                                & SECMR~\cite{SECMR_ding2023semantic}                           & Hierarchical Semantic Exploitation                                                                                                               & MIMIC-CXR~\cite{MIMIC-CXR_johnson2019mimic}               \\ \cmidrule{2-4} 
                                & DMACH~\cite{DMCAH_zhang2022deep}                           & Recurrent Attention for Local Feature Capture                                                                                                    & MIMIC-CXR~\cite{MIMIC-CXR_johnson2019mimic}               \\ \hline
\multirow{11}{*}{Real-valued}    & MMDL~\cite{MMDL_yu2021multimodal}                            & Customizable Modality (Image or Image-Text) Query                                                                                                & MIMIC-CXR~\cite{MIMIC-CXR_johnson2019mimic}               \\ \cmidrule{2-4} 
                                & \multicolumn{1}{c|}{\cite{mbilinyi2021retrieving}} & Image-Text Query, Deep Metric Learning via Triplet Loss                                                                                          & CheXpert~\cite{chexpert_irvin2019chexpert}                \\ \cmidrule{2-4} 
                                & ~\cite{ecg_qiu2023automated}                             & ECG Signals to Visual Representation Conversion                                                                                                  & PTB-XL~\cite{ptb-xl_wagner2020ptb}                  \\ \cmidrule{2-4} 
                                & \multirow{3}{*}{X-TRA~\cite{X-TRA_van2023x}}          & \multirow{3}{*}{\begin{tabular}[c]{@{}c@{}}CLIP Framework Alignment\\ Retrieval Augmentation\end{tabular}}                                       & MIMIC-CXR~\cite{MIMIC-CXR_johnson2019mimic}               \\
                                &                                 &                                                                                                                                                  & OpenI~\cite{iu_xray_demner2016preparing}                   \\
                                &                                 &                                                                                                                                                  & CheXpert~\cite{chexpert_irvin2019chexpert}                \\ \cmidrule{2-4} 
                                & Text-guided~\cite{text-guided_serieys2022text}                     & CLIP Framework Alignment                                                                                                                         & ROCO~\cite{roco_pelka2018radiology}                    \\ \cmidrule{2-4} 
                                & MCR~\cite{MCR_wei2023masked}                             & Masked Contrastive Reconstruction                                                                                                                & MIMIC-CXR~\cite{MIMIC-CXR_johnson2019mimic}               \\ \cmidrule{2-4} 
                                & MedFinder~\cite{chen2024bimcv}                       & 3D Volume Retrieval                                                                                                                              & BIMCV-R~\cite{chen2024bimcv}                 \\ \cmidrule{2-4} 
                                & \multirow{2}{*}{3D-MIR~\cite{abacha20233d}}         & \multirow{2}{*}{\begin{tabular}[c]{@{}c@{}}4 Different Aatomies \\ Support Slice-based, Volume-based, and Multimodal Queries\end{tabular}} & \multirow{2}{*}{3D-MIR~\cite{abacha20233d}} \\
                                &                                 &                                                                                                                                                  &                         \\ \hline
\end{tabular}
}
\end{table*}

As shown in Figure \ref{fig:retrieval}, image-text retrieval comprises two phases. In the initial training phase, modality encoders are trained to project data from different modalities into a common feature space. Subsequently, during the retrieval phase, an offline indexation stage occurs where the trained encoders extract representations and store them in a shared feature database. During the online search phase, real-time representations of query data are obtained, and data retrieval and ranking are conducted based on the similarity scores of their respective representations \cite{text-guided_serieys2022text}.
Concretely, the dual-encoder is extensively employed in medical image-text retrieval \cite{SECMR_ding2023semantic, DMCAH_zhang2022deep, text-guided_serieys2022text, mbilinyi2021retrieving, X-TRA_van2023x, category_zhang2022category, MultiModal_Medical_zhang2023multi}. In this design, each modality undergoes encoding through its respective modality-specific encoder. Nonetheless, to address the samples with missing modalities, MMDL \cite{MMDL_yu2021multimodal} foregoes the use of modality-specific encoders. Instead, it adopts a shared parameter strategy across the encoders to achieve modality invariance.
\xie{In particular, representation learning is essential for bridging different modalities in image-text retrieval. It maps image and text data into a common feature space by capturing their complementary information while minimizing redundancies. Current works for this representation can be divided into two categories (see Table \ref{tab:retrieval}): (1) binary representation learning, and (2) real-valued representation learning.}

\subsubsection{Binary Representation Learning}
Binary value representation learning is also known as cross-modal hash learning, where the encodings from different modalities are mapped into a shared Hamming space. The representations in hashing approaches are denoted as $h(x)\in\{-1,+1\}^{k}$, where $x$ is an instance from any modality and $k$ is the number of classes or clusters \cite{DMCAH_zhang2022deep}. In retrieval tasks, databases often store a massive number of features, necessitating substantial computational efforts for similarity computations.

Thus, deep hashing methods use binary encodings to facilitate the retrieval of similar instances through XOR operations instead of computing cosine similarities. In an XOR operation, identical bits yield a result of 0, while differing bits yield a result of 1. Consequently, the Hamming distance is the sum of the XOR results. This property enhances the computational and storage efficiency of hashing methods in this context~\cite{itr_cao2022imagetext}. 

\xie{Medical cross-modal hashing retrieval methods vary in their approaches to learning hash codes across modalities. SECMR \cite{SECMR_ding2023semantic} uses supervised training with class or pair labels, enhanced by leveraging hierarchical disease label associations to impose semantic constraints. 
Similarly, \cite{category_zhang2022category} employs category hash codes generated by a category hashing network to supervise multimodal hashing, and bridges the semantic gap between modalities using a union hash network to learn correlations between their hash codes.
DMACH \cite{DMCAH_zhang2022deep} focuses on cross-modal instance retrieval by combining global features, extracted via global average pooling, with fine-grained local features, captured by recurrent attention mechanisms, and projecting these into a common Hamming space.
}

\subsubsection{Real-valued Representation Learning}

\xie{Hashing methods improve retrieval time efficiency but at the cost of reduced accuracy due to simplified binary feature representation compared to real-value representations \cite{itr_cao2022imagetext}. Thus, some works \cite{MMDL_yu2021multimodal, text-guided_serieys2022text, MultiModal_Medical_zhang2023multi, mbilinyi2021retrieving, X-TRA_van2023x, ecg_qiu2023automated} use real-value representations for both image and text modalities.

MMDL \cite{MMDL_yu2021multimodal} uses Selective Convolution Descriptor Aggregation (SCDA) to identify abnormal regions as image inputs and projects image and text into a shared space with triplet loss. 
Similarly, \cite{mbilinyi2021retrieving} employs a Gated Multimodal Unit (GMU) and triplet loss for common space learning, supporting multimodal image-text queries and improving retrieval performance.
To further improve cross-modal retrieval, some methods use multi-level learning frameworks, such as \cite{MultiModal_Medical_zhang2023multi}, which leverages inter- and intra-modal associations to achieve more accurate retrieval results.

Recent studies focus on adapting pre-trained models for medical applications. \cite{text-guided_serieys2022text} adapts CLIP \cite{CLIP_radford2021learning} for the medical domain using contrastive learning to align image and text representations in a shared space. 
X-TRA \cite{X-TRA_van2023x} also builds on CLIP by adding a classification task to improve content-based feature extraction, with data indexed by FAISS \cite{FAISS_johnson2017billionscale} for multimodal retrieval. 
MCR \cite{MCR_wei2023masked} employs multiple pre-training tasks (Image-Text Contrastive Learning (ITC), Masked Image Modeling (MIM), and Masked Language Modeling (MLM)) and uses masked inputs to enhance task integration and reduce information interference, optimizing pre-training efficiency.

Certain methods are tailored to specific medical domains. For instance, \cite{ecg_qiu2023automated} addresses the complexity of ECG disease diagnosis by framing it as a retrieval problem involving past reports. In this approach, ECG signals are preprocessed into a visual format and treated as images, which are then jointly trained with corresponding reports to establish vision-language alignment. This alignment is achieved through three different pre-training objectives: ITC, Image-Text Matching (ITM), and Cross-modal MLM.
}

\noindent\textbf{3D Applications.}
MedFinder~\cite{chen2024bimcv} extends beyond traditional 2D medical image retrieval by targeting cross-modal retrieval for 3D images. It processes two augmented 3D images through a 3D ViT and applies MSE loss to their representations. The text stream uses a pre-trained BiomedCLIP encoder, with ITC loss applied between the text representation and a fused image representation derived from the augmented images. 3D-MIR~\cite{abacha20233d} offers a versatile multi-modal retrieval method that covers four anatomies: liver, colon, pancreas, and lung. It supports various query types, including 2D slices, 3D volumes, and multi-modal queries (image-text embeddings). The study finds that 3D volume-based retrieval is more effective for broad categorization, while slice-based retrieval is better at capturing fine-grained details.

\subsection{Limitation and Insights}

Medical image-text retrieval tasks still face several limitations that impede their effectiveness. In the feature extraction stage, capturing fine-grained features remains a significant challenge in the medical domain. The global semantics in medical images and reports often display similarities across diverse patients, whereas the subtle visual abnormalities and specific disease names offer improved discriminative potential. Therefore, it is essential to focus on extracting fine-grained features and leveraging these detailed features in retrieval processes to improve accuracy and relevance.
\xie{Besides, there is a pressing need for computationally efficient retrieval methods capable of managing large-scale datasets and complex queries without sacrificing performance. FAISS~\cite{FAISS_johnson2017billionscale} is a significant step forward by offering a powerful library for performing similarity searches and clustering on dense vectors with high efficiency. However, there remains a need to develop more advanced techniques that can seamlessly integrate into existing clinical workflows.}
Moreover, many works remain at the proof-of-concept stage, with limited validation in real-world clinical environments. This lack of extensive real-world testing restricts their practical applicability and robustness, underscoring the need for further clinical validation to ensure they meet the demands of actual clinical use.

\section{Challenges and Potential Future Directions}\label{challenges}
\subsection{Medical Data}

\noindent\textbf{Limited Data.} Medical data is scarcer compared to natural data. They must be obtained from real medical cases, which are private and owned by the patients. There are numerous ethical, legal, and privacy procedures to access medical data, making it more challenging to train an accurate medical vision-language model. Besides, these limited datasets cover only a small part of the medical domain. When dealing with rare diseases, there is not enough data available, causing the inability to train accurate medical vision-language models.
\qinew{With the growing of data generation methods, \eg, diffusion models~\cite{ho2020denoising,rombach2022high}, generating multi-modal medical data~\cite{xie2024pairaug} has become a promising avenue for overcoming challenges of data scarcity. However, effectively adapting these models to the medical domain remains an open question.}

\noindent\textbf{Unpaired Image-Text Data.} The data collection process can encounter mismatches between medical images and texts, which may arise from various issues, including annotation errors, asynchronous timing between imaging and report generation, and confusion between multiple imaging types or body parts. Data loss or corruption can also occur during data conversion or storage, adding another layer of complexity to developing robust medical vision-language models.

\noindent\textbf{Medical Data \vs~Natural Data.} 
(1) Medical images have a high resolution, but the structures or abnormalities that models need to focus on are often tiny and may even be only a few pixels. They also tend to contain a great deal of medical information in relatively small areas, \eg, a CT scan image may show tiny structures in the brain or subtle changes in a tumor. 
These localized structures in images often have important biological and clinical significance and therefore require a high degree of attention.
In contrast, natural images contain objects at a variety of scales, ranging from microscopic to macroscopic coverage. 
The specificity of medical images poses a challenge in building medical models. They are often affected by noise and distracting patterns, which makes it more difficult to identify diagnostic information. 
%
\qi{
(2) Modalities of medical data are more complex such as X-rays, MRIs, CT scans, pathology slides, and reports. Each modality captures different aspects of a patient’s condition, and the way these modalities are used and processed can vary, creating a gap when integrating multimodal information. For instance, combining data from a CT scan with a textual diagnosis report requires handling both image-specific and language-specific features.
(3) Medical data is collected from a wide variety of sources, including different hospitals, imaging devices, and medical practices. This variation leads to differences in the quality, format, and style of data, which can introduce inconsistencies when training models. For example, radiology images from one institution may vary in resolution, contrast, and noise levels compared to another.
}

\subsection{Medical Vision-Language Alignment}

Medical vision-language models rely on aligning text data (like clinical reports and medical literature) with image data (like medical images and pathology slides) for effective cross-modal learning. However, semantic differences between medical texts and images, such as the complex mapping between disease descriptions and image features, present challenges. Medical texts are also uniquely complex, containing varied information like diagnostic findings, terminology, and patient details, requiring advanced language processing. Besides, medical language is highly specialized, using domain-specific terms, abbreviations, and unique syntax. Accurate alignment with visual features is essential yet non-trivial for building effective medical vision-language models.

\subsection{Evaluation Metrics}

Existing natural language evaluation metrics are limited when directly applied to medical tasks due to the complexity of medical concepts and the need for subjective expert assessments, which are time-consuming. Evaluating medical vision-language models is further complicated by the need to understand interactions between modalities and assess multiple tasks simultaneously, such as disease diagnosis, image analysis, and text comprehension. Besides, the imbalance in medical data, with some diseases being rare, poses challenges. Evaluation metrics must reflect the model's usefulness and performance in clinical practice, focusing on accuracy, completeness, and consistency with clinical reality, rather than just syntactic correctness or fluency. Moreover, metrics must account for subjective differences among experts, ensuring that evaluations accurately represent real-world clinical practices.

\subsection{Efficiency and Clinical Implementation}
\qi{
For medical vision-language models to be adopted in clinical practice, they should balance performance with computational efficiency. For example, (near) real-time image-to-report generation is critical for reducing workload and enhancing decision-making speed in busy clinical environments. Models need to deliver accurate results while being computationally lightweight to ensure smooth integration into existing clinical workflows, which often have limited processing resources. Besides, scalability and ease of use are vital to facilitate widespread implementation across various healthcare settings. Developing efficient models with minimal latency and resource requirements is key to their successful clinical adoption.
}

\subsection{Reliability and Interpretability}
\qi{
In medical applications, reliability and interpretability are paramount for ensuring trust and safety in vision-language models. Clinicians must be able to rely on the model's outputs for accurate diagnoses and treatment decisions. This requires models to not only be highly accurate but also explainable—offering clear reasoning for their decisions so medical professionals can understand how conclusions are drawn. Interpretability enhances the model's transparency, allowing clinicians to verify its outputs against medical knowledge. Ensuring reliability and interpretability is essential for fostering trust among healthcare providers and for facilitating the model's integration into clinical practice.
}

\section{Conclusion}\label{conclusion}
In summary, our survey introduces the emerging field of vision-language in medicine, shows the advantages and existing achievements of vision-language modeling for medical tasks, and highlights its potential to significantly improve various aspects of healthcare services. We have selected medical report generation, medical visual question answering, medical multimodal diagnosis and prognosis, medical image segmentation, and medical image-text retrieval as representative studies in the medical domain. We present the broad process of each task and analyze in depth the challenges and difficulties these tasks face. By investigating a large number of studies and papers from recent years, we categorize and summarize all these recent model designs in a clear and concise form. Finally, we also point out the shortcomings of the current model as well as identify future research directions. 

By analyzing the key tasks, we have witnessed improvements and innovations in techniques for aligning and fusing visual and textual information.
While significant progress has been made in vision-language modeling to assist the medical decision-making process, multiple future directions have potential and should be further explored. 
The first direction is to further overcome the limitations imposed by data scarcity. By generating a wider range of datasets or more efficient data augmentation methods, models can be regularized with richer and more diverse data resources.
Second, future research can continue to develop methods and techniques that can effectively fuse different data modalities to improve the performance and application scope of MVLMs. With the development of LLMs, the ability of MVLMs to understand text will be further improved.
The last direction is to improve the generalization performance and interpretability of the model. It should not only be applicable to different medical fields and various practical application scenarios but also explain the methods and model outcomes, helping medical practitioners and clinical researchers better understand the decision-making process and results of the model.


\bmhead{Data availability statements} The data supporting the findings of this study are
openly available, as summarized in the Appendix. Our open-sourced repository includes the full collection of papers and code of this work.

{
    \small
    \bibliography{sn-bibliography}
}


\clearpage

\begin{appendices}

\section{Supplementary Material}

\subsection{Datasets and Evaluation Metrics in Medical Report Generation}\label{mrg_dataset}

\subsubsection{Datasets}

\noindent\textbf{IU X-ray} \cite{iu_xray_demner2016preparing}: The IU X-ray contains 7470 images and 3955 reports. Each report consists of the following sections: impression, findings, tags, comparison, and indication.

\noindent\textbf{MIMIC-CXR} \cite{MIMIC-CXR_johnson2019mimic}: MIMIC-CXR contains 377,110 chest X-rays paired with 227,835 associated radiology reports from 64,588 patients of the Beth Israel Deaconess Medical Center.

\noindent\textbf{PadChest} \cite{padchest_bustos2020padchest}: PadChest includes 160,868 images obtained from 67,625 patients that were interpreted and reported by radiologists at Hospital San Juan Hospital (Spain) from 2009 to 2017, covering six different position views and additional information on image acquisition and patient demography.

\noindent\textbf{CheXpert} \cite{chexpert_irvin2019chexpert}: CheXpert contains 224,316 chest radiographs of 65,240 patients with both frontal and lateral views available. CheXpert labels the presence of 14 observations in radiology reports including no finding, enlarged cardiom, cardiomegaly, lung lesion, lung opacity, edema, consolidation, pneumonia, atelectasis, pneumothorax, pleural effusion, pleural other, fracture and support devices. 

\noindent\textbf{ROCO} \cite{roco_pelka2018radiology}: ROCO is stratified into two categories: radiology and out-of-class. The radiology group includes 81,825 radiology images, including computer tomography (CT), ultrasound, x-ray, fluoroscopy, positron emission tomography (PET), mammography, magnetic resonance imaging (MRI), angiography, and PET-CT. The out-of-class group has 6,127 images, including synthetic radiology images, clinical photos, portraits, compound radiology images, and digital art.

In addition to the datasets above, the following datasets are also used for the medical report generation task: ImageCLEF Caption 2017 \cite{eickhoff2017overview}, ImageCLEF Caption 2018 \cite{garcia2018overview}, INBreast \cite{moreira2012inbreast}, STARE, COVID-CTR \cite{li2023auxiliary}, JLiverCT \cite{nishino2022factual}, Japanese Computed Tomography (JCT), FFA-IR \cite{li2021ffa}, SD \cite{allan20202018}.
Experimental results of existing works on the IU and MIMIC datasets are displayed in Tables~\ref{IU} and~\ref{MIMIC}, respectively.

\begin{table*}[htbp]
  \renewcommand\arraystretch{1.25}
  \belowrulesep=0pt
  \aboverulesep=0pt
    \centering
    \caption{Results of existing medical report generation methods on the IU dataset.}
    \resizebox{\linewidth}{!}{
      \begin{tabular}{c|ccccccccc}
      \toprule
      Methods & Source & Highlight & B1 & B2 & B3 & B4 & ROUGE & MTOR & CIDEr  \\
      \midrule
      \multirow{2}*{KGAE \cite{liu2021auto}} & \multirow{2}*{NIPS 2021} & Pre-constructed knowledge graph & \multirow{2}*{0.512} & \multirow{2}*{0.327} & \multirow{2}*{0.240} & \multirow{2}*{0.179} & \multirow{2}*{0.383} & \multirow{2}*{0.195} & \multirow{2}*{-} \\
      & & Knowledge-driven encoder and decoder & & & & & & & \\
      \multirow{2}*{MedWriter \cite{yang2021writing}} & \multirow{2}*{ACL 2021} & Memory retrieval & \multirow{2}*{0.471} & \multirow{2}*{0.336} & \multirow{2}*{0.238} & \multirow{2}*{0.166} & \multirow{2}*{0.382} & \multirow{2}*{-} & \multirow{2}*{0.345}  \\
      & & Multi-query attention & & & & & & & \\
      CMCL \cite{liu2022competence} & ACL 2021 & Competence-based multimodal curriculum learning & 0.473 & 0.305 & 0.217 & 0.162 & 0.378 & 0.186 & -  \\
      CA \cite{liu2021contrastive} & ACL 2021 & Contrastive Attention & 0.492 & 0.314 & 0.222 & 0.169 & 0.381 & 0.193 & -  \\
      \multirow{2}*{CMN \cite{chen2022cross}} & \multirow{2}*{ACL 2021} & Cross-modal memory networks & \multirow{2}*{0.475} & \multirow{2}*{0.309} & \multirow{2}*{0.222} & \multirow{2}*{0.170} & \multirow{2}*{0.375} & \multirow{2}*{0.191} & \multirow{2}*{-} \\
      & & Memory matrix & & & & & & & \\
      MEDSKIP \cite{pahwa2021medskip} & \multirow{2}*{ICCV 2021} & Modified HRNet including added skip connections & 0.467 & 0.297 & 0.214 & 0.162 & 0.355 & 0.187 & - \\
      MEDSKIP+CBAM \cite{pahwa2021medskip} & & along with convolutional block attention & 0.467 & 0.303 & 0.210 & 0.155 & 0.371 & 0.197 & - \\
      PPKED \cite{liu2021exploring} & CVPR 2021 & Posterior-and-Prior Knowledge Exploring-and-Distilling & 0.483 & 0.315 & 0.224 & 0.168 & 0.376 & 0.190 & 0.351 \\
      \multirow{2}*{AlignTransformer \cite{you2021aligntransformer}} & \multirow{2}*{MICCAI 2021} & Multi-grained visual features & \multirow{2}*{0.484} & \multirow{2}*{0.313} & \multirow{2}*{0.225} & \multirow{2}*{0.173} & \multirow{2}*{0.379} & \multirow{2}*{0.204} & \multirow{2}*{-} \\
      & & Hierarchically alignment & & & & & & & \\
      \multirow{2}*{$M^2$ TR.PROGRESSIVE} & \multirow{2}*{EMNLP 2021} & Progressive text generation & \multirow{2}*{0.486} & \multirow{2}*{0.317} & \multirow{2}*{0.232} & \multirow{2}*{0.173} & \multirow{2}*{0.390} & \multirow{2}*{0.192} & \multirow{2}*{-} \\
      \cite{nooralahzadeh2021progressive} & & incorporating high-level concepts & & & & & & & \\
      \multirow{2}*{MV+T+I \cite{nguyen2021automated}} & \multirow{2}*{EMNLP 2021} & Context modeling and disease-state aware & \multirow{2}*{0.515} & \multirow{2}*{0.378} & \multirow{2}*{0.293} & \multirow{2}*{0.235} & \multirow{2}*{0.436} & \multirow{2}*{0.219} & \multirow{2}*{-} \\
      & & Consistency enhancement & & & & & & & \\
      VTI-LSTM \cite{najdenkoska2022uncertainty} & \multirow{2}*{MIA 2022} & \multirow{2}*{Variational topic inference} & 0.493 & 0.360 & 0.291 & 0.154 & 0.375 & 0.218 & -  \\
      VTI-TRS \cite{najdenkoska2022uncertainty} & & & 0.503 & 0.394 & 0.302 & 0.170 & 0.390 & 0.230 & - \\
      TranSQ \cite{kong2022transq} & MICCAI 2022 & Sentence retrieval and selection & 0.484 & 0.333 & 0.238 & 0.175 & 0.415 & 0.207 & - \\
      \multirow{2}*{SGF \cite{li2022self}} & \multirow{2}*{MICCAI 2022} & Suite of unsupervised and supervised deep learning & \multirow{2}*{0.467} & \multirow{2}*{0.334} & \multirow{2}*{0.261} & \multirow{2}*{0.215} & \multirow{2}*{0.415} & \multirow{2}*{0.201} & \multirow{2}*{-} \\
      & & Similarity comparison mechanism & & & & & & & \\
      \multirow{2}*{ITA \cite{wang2022inclusive}} & \multirow{2}*{MICCAI 2022} & Inclusive Task-Aware Framework & \multirow{2}*{0.505} & \multirow{2}*{0.340} & \multirow{2}*{0.247} & \multirow{2}*{0.188} & \multirow{2}*{0.382} & \multirow{2}*{0.208} & \multirow{2}*{-} \\
      & & Classification token+auto-balance mask loss & & & & & & & \\
      Repsnet \cite{tanwani2022repsnet} & MICCAI 2022 & Contrastive learning & 0.580 & 0.440 & 0.320 & 0.270 & - & - & - \\
      \multirow{2}*{XPRONET \cite{wang2022cross}} & \multirow{2}*{ECCV 2022} & End-to-end cross-modal prototype driven network & \multirow{2}*{0.525} & \multirow{2}*{0.357} & \multirow{2}*{0.262} & \multirow{2}*{0.199} & \multirow{2}*{0.411} & \multirow{2}*{0.220} & \multirow{2}*{0.359} \\
      & & Improved multi-label contrastive loss & & & & & & & \\
      \cite{yan2022prior} & JBHI 2022 & Prior Guided Attention+sparse attention & 0.482 & 0.313 & 0.232 & 0.181 & 0.381 & 0.203 & 0.735  \\
      MMTN \cite{cao2023mmtn} & AAAI 2023 & Multi-modal Memory Transformer Network & 0.486 & 0.321 & 0.232 & 0.175 & 0.375 & - & 0.361  \\
      ATAG \cite{yan2023attributed} & TMI 2023 & Attributed abnormality graph+gating mechanism & \multicolumn{4}{c}{0.256 (Average of B1$\sim$B4)} & 0.341 & - & 0.380  \\
      ICT \cite{zhang2023novel} &JBHI 2023 & Inter-intra feature extraction & 0.5001 & 0.3395 & 0.2484 & 0.1897 & 0.3942 & 0.2069 & - \\
      \multirow{2}*{EDC-Net \cite{singh2022efficient}} & \multirow{2}*{JBHI 2023} & Deep ensemble network & \multirow{2}*{0.516} & \multirow{2}*{0.348} & \multirow{2}*{0.238} & \multirow{2}*{0.178} & \multirow{2}*{-} & \multirow{2}*{-} & \multirow{2}*{-} \\
      & & Self-adaptive parameter control-based differential evolution & & & & & & & \\
      KiUT \cite{huang2023kiut} & CVPR 2023 & Symptom graph & 0.525 & 0.360 & 0.251 & 0.185 & 0.409 & 0.242 & - \\
      DCL \cite{li2023dynamic} & CVPR 2023 & Dynamic graph with contrastive learning paradigm & - & - & - & 0.163 & 0.383 & 0.193 & 0.586  \\
      METransformer \cite{wang2023metransformer} & CVPR 2023 & Multi-expert joint diagnosis & 0.483 & 0.322 & 0.228 & 0.172 & 0.380 & 0.192 & 0.435  \\
      \bottomrule
      \end{tabular}%
      }
    \label{IU}%
  \end{table*}%

\begin{table*}[htbp]
  \renewcommand\arraystretch{1.25}
  \belowrulesep=0pt
  \aboverulesep=0pt
    \centering
    \caption{Results of existing medical report generation methods on the MIMIC dataset.}
    \resizebox{\linewidth}{!}{
      \begin{tabular}{c|ccccccccc}
      \toprule
      Methods & Source & Highlight & B1    & B2    & B3    & B4    & ROUGE & MTOR  & CIDEr \\
      \midrule
      \multirow{2}*{KGAE \cite{liu2021auto}} & \multirow{2}*{NIPS 2021} & Pre-constructed knowledge graph & \multirow{2}*{0.369} & \multirow{2}*{0.231} & \multirow{2}*{0.156} & \multirow{2}*{0.118} & \multirow{2}*{0.295} & \multirow{2}*{0.153} & \multirow{2}*{-} \\
      & & Knowledge-driven encoder and decoder & & & & & & & \\
      \multirow{2}*{\cite{zhou2021visual}} & \multirow{2}*{ICCV 2021} & Multi-modality semantic attention & \multirow{2}*{0.372} & \multirow{2}*{0.241} & \multirow{2}*{0.168} & \multirow{2}*{0.123} & \multirow{2}*{0.335} & \multirow{2}*{0.190} & \multirow{2}*{1.121} \\
      & & Topic-level losses & & & & & & & \\
      PPKED \cite{liu2021exploring} & CVPR 2021 & Posterior-and-Prior Knowledge Exploring-and-Distilling & 0.360 & 0.224 & 0.149 & 0.106 & 0.284 & 0.149 & 0.237 \\
      \cite{yang2021joint} & TMM 2021 & Triple-branch encoding+soft attention mechanism & 0.362 & 0.251 & 0.188 & 0.143 & 0.326 & - & 0.273 \\
      \multirow{2}*{MV+T+I \cite{nguyen2021automated}} & \multirow{2}*{EMNLP 2021} & Context modeling and disease-state aware mechanism & \multirow{2}*{0.495} & \multirow{2}*{0.360} & \multirow{2}*{0.278} & \multirow{2}*{0.224} & \multirow{2}*{0.390} & \multirow{2}*{0.222} & \multirow{2}*{-} \\
      & & Consistency enhancement & & & & & & & \\
      CMCL \cite{liu2022competence} & ACL 2021 & Competence-based multimodal curriculum learning & 0.344 & 0.217 & 0.140 & 0.097 & 0.281 & 0.133 & - \\
      \multirow{2}*{MedWriter \cite{yang2021writing}} & \multirow{2}*{ACL 2021} & Memory retrieval & \multirow{2}*{0.438} & \multirow{2}*{0.297} & \multirow{2}*{0.216} & \multirow{2}*{0.164} & \multirow{2}*{0.332} & \multirow{2}*{-} & \multirow{2}*{0.306} \\
      & & Multi-query attention & & & & & & & \\
      CA \cite{liu2021contrastive} & ACL 2021 & Contrastive Attention & 0.350 & 0.219 & 0.152 & 0.109 & 0.283 & 0.151 & -\\
      \multirow{2}*{CMN \cite{chen2022cross}} & \multirow{2}*{ACL 2021} & Cross-modal memory networks & \multirow{2}*{0.353} & \multirow{2}*{0.218} & \multirow{2}*{0.148} & \multirow{2}*{0.106} & \multirow{2}*{0.278} & \multirow{2}*{0.142} & \multirow{2}*{-} \\
      & & Memory matrix & & & & & & & \\
      AlignTransformer \cite{you2021aligntransformer} & MICCAI 2021 & Multi-grained visual features+hierarchically alignment & 0.378 & 0.235 & 0.156 & 0.112 & 0.283 & 0.158 & - \\
      RATCHET \cite{hou2021ratchet} & MICCAI 2021 & Transformer-based CNN-RNN & 0.232  & - & - & - & 0.240 & 0.101 & 0.493 \\
      \cite{yan2022prior} & JBHI 2022 & Prior Guided Attention+sparse attention & 0.356 & 0.222 & 0.151 & 0.111 & 0.280 & 0.140 & 0.154 \\
      \multirow{2}*{XPRONET \cite{wang2022cross}} & \multirow{2}*{ECCV 2022} & End-to-end cross-modal prototype driven network & \multirow{2}*{0.344} & \multirow{2}*{0.215} & \multirow{2}*{0.146} & \multirow{2}*{0.105} & \multirow{2}*{0.279} & \multirow{2}*{0.138} & \multirow{2}*{-} \\
      & & Improved multi-label contrastive loss & & & & & & & \\
      TranSQ \cite{kong2022transq} & MICCAI 2022 & Sentence retrieval and selection & 0.423 & 0.261 & 0.171 & 0.116 & 0.286 & 0.168 & - \\
      \multirow{2}*{MCGN \cite{wang2022medical}} & \multirow{2}*{MICCAI 2022} & Bilinear-pooling-assisted sparse attention block & \multirow{2}*{0.413} & \multirow{2}*{0.266} & \multirow{2}*{0.186} & \multirow{2}*{0.136} & \multirow{2}*{0.298} & \multirow{2}*{0.170} & \multirow{2}*{0.429} \\
      & & Medical concepts generation network & & & & & & & \\
      \multirow{2}*{ITA \cite{wang2022inclusive}} & \multirow{2}*{MICCAI 2022} & Inclusive Task-Aware Framework & \multirow{2}*{0.395} & \multirow{2}*{0.253} & \multirow{2}*{0.170} & \multirow{2}*{0.121} & \multirow{2}*{0.284} & \multirow{2}*{0.147} & \multirow{2}*{-} \\
      & & Classification token+auto-balance mask loss & & & & & & & \\
      \cite{yang2022knowledge} & MIA 2022 & Knowledge-enhanced multi-head attention & 0.363 & 0.228 & 0.156 & 0.115 & 0.284 & - & 0.203 \\
      VTI-LSTM \cite{najdenkoska2022uncertainty} & \multirow{2}*{MIA 2022} & \multirow{2}*{Variational topic inference} & 0.418 & 0.293 & 0.152 & 0.109 & 0.302 & 0.177 & - \\
      VTI-TRS \cite{najdenkoska2022uncertainty} & & & 0.475 & 0.314 & 0.196 & 0.136 & 0.315 & 0.191 & - \\
      MMTN \cite{cao2023mmtn} & AAAI 2023 & Multi-modal Memory Transformer Network & 0.379 & 0.238 & 0.159 & 0.116 & 0.283 & 0.161 & - \\
      ICT \cite{zhang2023novel} & JBHI 2023 & Inter-intra feature extraction & 0.3734 & 0.2303 & 0.1559 & 0.1125 & 0.2759 & 0.1430 & - \\
      \multirow{2}*{\cite{dalla2023finding}} & \multirow{2}*{MICCAI 2023} & Multi-task Faster R-CNN & \multirow{2}*{0.490} & \multirow{2}*{0.363} & \multirow{2}*{0.288} & \multirow{2}*{0.237} & \multirow{2}*{0.406} & \multirow{2}*{0.213} & \multirow{2}*{-} \\
      & & Triples representation & & & & & & & \\
      KiUT \cite{huang2023kiut} & CVPR 2023 & Symptom graph & 0.393 & 0.243 & 0.159 & 0.113 & 0.285 & 0.160 & - \\
      DCL \cite{li2023dynamic} & CVPR 2023 & Dynamic graph with contrastive learning & - & - & - & 0.109 & 0.284 & 0.150 & 0.281 \\
      RGRG \cite{tanida2023interactive} & CVPR 2023 & Region-guided report generation & 0.373 & 0.249 & 0.175 & 0.126 & 0.264 & 0.168 & 0.495 \\
      METransformer \cite{wang2023metransformer} & CVPR 2023 & Multi-expert joint diagnosis mechanism & 0.386 & 0.250 & 0.169 & 0.124  & 0.291 & 0.152 & 0.362 \\
      \multirow{2}*{\cite{yang2023radiology}} & \multirow{2}*{MIA 2023} & Novel knowledge updating mechanism & \multirow{2}*{0.386} & \multirow{2}*{0.237} & \multirow{2}*{0.157} & \multirow{2}*{0.111} & \multirow{2}*{0.274} & \multirow{2}*{-} & \multirow{2}*{0.111} \\
      & & Multi-modal alignment & & & & & & & \\
      \bottomrule
      \end{tabular}%
      }
    \label{MIMIC}%
  \end{table*}%

\subsubsection{Natural Language Evaluation Metrics}
\noindent\textbf{BLEU} (Bilingual evaluation understudy) \cite{papineni2002bleu} measures the similarity between the generated report and the ground truth report by calculating the overlap of the word n-grams. BLEU considers n-grams of different lengths, including BLEU-1, BLEU-2, BLEU-3 and BLEU-4. These n-gram matches are weighted equally, but different lengths of n-grams contribute differently to the translation quality.

\noindent\textbf{METEOR} (Metric for Evaluation of Translation with Explicit ORdering) \cite{banerjee2005meteor} is an extension of BLEU-1 that performs unigram matching of generated text with all the reference text. METEOR employs an F-score by taking a harmonic mean of unigram precision and recall with a bias towards recall.

\noindent\textbf{ROUGE-L} (Recall-Oriented Understudy for Gisting Evaluation) \cite{lin2004rouge} is a recall-based metric, originally developed for the evaluation of text summarization algorithms. Recall is the ratio of matched n-grams out of the total number of n-grams in the reference text.

\noindent\textbf{CIDEr} (Consensus-based Image Description Evaluation) 
\cite{vedantam2015cider} measures the consensus by calculating the cosine similarity between n-gram TF–IDF representations of the generated report and the reference report.

\subsubsection{Clinical Efficacy Metrics}

\noindent\textbf{Medical Subject Heading (MeSH) Accuracy (MA)} \cite{huang2019multi} is the ratio of the number of MeSH terms correctly generated by a model to the number of all MeSH terms in the reports of the OpenI dataset.

\noindent\textbf{Clinical Efficacy (CE)} \cite{liu2019clinically} measures the accuracy, precision, and recall of disease labels extracted by CheXpert from the ground truth references and the generated reports.

\noindent\textbf{Medical Image Report Quality Index (MIRQI)} \cite{zhang2020radiology} evaluates the quality of 
paired reports by graph matching.

\noindent\textbf{Factually-oriented} \cite{delbrouck2022improving} measures the factual correctness and completeness of the generated radiology reports.

\noindent\textbf{CO} \cite{nishino2022factual} quantifies the consistency of the description order of the generated reports.

\noindent\textbf{nTKD} \cite{zhou2021visual} evaluates whether the sentences in the generated report include all identified diseases and their detailed descriptive details.

\subsection{Datasets and Evaluation Metrics in Medical Visual Question Answering}\label{vqa_dataset}
\subsubsection{Datasets}

\begin{table*}[htbp]
\belowrulesep=0pt
\aboverulesep=0pt
  \centering
  \caption{A brief introduction of commonly used VQA datasets. ``Synthetical'' means that the questions and answers have been filtered or annotated by experts.}
  \resizebox{\linewidth}{!}{
    \begin{tabular}{c|ccccl}
    \toprule
    \textcolor[rgb]{ 0,  .439,  .753}{\textbf{Dataset}} & \textcolor[rgb]{ 0,  .439,  .753}{\textbf{\#Images}} & \textcolor[rgb]{ 0,  .439,  .753}{\textbf{\#QA pairs}} & \multicolumn{1}{l}{\textcolor[rgb]{ 0,  .439,  .753}{\textbf{Source of images and content}}} & \multicolumn{1}{l}{\textcolor[rgb]{ 0,  .439,  .753}{\textbf{QA Creation}}} & \textcolor[rgb]{ 0,  .439,  .753}{\textbf{Question Category}} \\
    \midrule
    VQA-Med 2018 & 2866  & 6413  & \multicolumn{1}{l}{PubMed Central Articles} & Synthetical & - Location \\
          &       &       &       &       & - Finding \\
          &       &       &       &       & - Yes/No questions \\
          &       &       &       &       & - Other questions \\
    \midrule
    VQA-Med 2019 & 4200  & 15292 & \multicolumn{1}{l}{MedPix database:} & Synthetical & - Modality \\
          &       &       & \multicolumn{1}{l}{- Various in 36 modalities, 16} &       & - Plane \\
          &       &       & \multicolumn{1}{l}{planes, and 10 organ systems} &       & - Organ System \\
          &       &       &       &       & - Abnormality \\
    \midrule
    VQA-Med 2020 & 5000  & 5000  & \multicolumn{1}{l}{MedPix database} & Synthetical & - Abnormality \\
    \midrule
    VQA-Med 2021 & 5000  & 5000  & \multicolumn{1}{l}{MedPix database} & Synthetical & - Abnormality \\
    \midrule
    VQA-RAD & 315   & 3515  & \multicolumn{1}{l}{MedPix database:} & Natural & - Modality \\
          &       &       & \multicolumn{1}{l}{- Head axial single-slice CTs or} &       & - Plane \\
          &       &       & \multicolumn{1}{l}{MRIs} &       & - Organ System \\
          &       &       & \multicolumn{1}{l}{- Chest X-rays} &       & - Abnormality \\
          &       &       & \multicolumn{1}{l}{- Abdominal axial CTs} &       & - Object/Condition Presence \\
          &       &       &       &       & - Positional reasoning \\
          &       &       &       &       & - Color \\
          &       &       &       &       & - Size \\
          &       &       &       &       & - Attribute Other \\
          &       &       &       &       & - Counting \\
          &       &       &       &       & - Other \\
    \midrule
    PathVQA & 4998  & 32799 & \multicolumn{1}{l}{Electronic pathology textbooks} & Synthetical & - Color \\
          &       &       & \multicolumn{1}{l}{PEIR Digital Library} &       & - Location \\
          &       &       &       &       & - Appearance \\
          &       &       &       &       & - Shape \\
          &       &       &       &       & - etc \\
    \midrule
    SLAKE & 642   & 14028 & \multicolumn{1}{l}{Medical Segmentation Decathlon,} & Natural & - Organ \\
          &       &       & \multicolumn{1}{l}{NIH Chest X-ray,} &       & - Position \\
          &       &       & \multicolumn{1}{l}{CHAOS:} &       & - Knowledge Graph \\
          &       &       & \multicolumn{1}{l}{- Chest X-rays/CTs} &       & - Abnormality \\
          &       &       & \multicolumn{1}{l}{- Abdomen CTs/MRIs} &       & - Modality \\
          &       &       & \multicolumn{1}{l}{- Head CTs/MRIs} &       & - Plane \\
          &       &       & \multicolumn{1}{l}{- Neck CTs} &       & - Quality \\
          &       &       & \multicolumn{1}{l}{-Pelvic cavity CTs} &       & - Color \\
          &       &       &       &       & - Size \\
          &       &       &       &       & - Shape \\
    \midrule
    IDRiD & 516   & 220k  & \multicolumn{1}{l}{Eye Clinic in India} & Synthetical & - Yes/No questions \\
    \midrule
    BACH  & 400+  & 360k  & \multicolumn{1}{l}{Annotated Dataset} & Synthetical & - Abnormality \\
          &       &       &       &       & - Size \\
          &       &       &       &       & - Counting \\
          &       &       &       &       & - Yes/No questions \\
    \midrule
    Tools & 2532  & 1M    & \multicolumn{1}{l}{Cholecystectomy Video} & Synthetical & - Position \\
          &       &       &       &       & - Counting \\
          &       &       &       &       & - Yes/No questions \\
    \bottomrule
    \end{tabular}%
    }
  \label{VQA_introduction}%
\end{table*}%

\noindent\textbf{VQA-Med 2018} \cite{hasan2018overview}: VQA-Med 2018 is the first publicly available dataset in the medical domain. The QA pairs were generated from captions by a semi-automatic approach. Two expert human annotators manually checked all generated QA pairs in two passes.

\noindent\textbf{VQA-Med 2019} \cite{abacha2019vqa}: VQA-Med 2019 includes a training set of 3200 medical images with 12,792 QA pairs, a validation set of 500 medical images with 2000 QA pairs and a test set of 500 medical images with 500 QA pairs.

\noindent\textbf{VQA-Med 2020}: VQA-Med 2020 is the third edition of the VQA-Med. The training set includes 4,000 radiology images with 4,000 associated QA pairs. The validation set includes 500 radiology images with 500 QA pairs. The test set includes 500 radiology images with 500 associated questions.

\noindent\textbf{VQA-Med 2021} \cite{ben2021overview}: VQA-Med 2021 is created under the principles as those in VQA-Med-2020. The training set is the same dataset as used in VQA-Med-2020. The validate set and test set are new and manually reviewed by medical doctors.

\noindent\textbf{VQA-RAD} \cite{lau2018dataset}: VQA-RAD contains 104 head axial single-slice CTs or MRIs, 107 chest x-rays, and 104 abdominal axial CTs. All these 315 images are labeled with the organs they are taken from. There are 3,515 questions with 11 different types in VQA-RAD.

\noindent\textbf{PathVQA} \cite{he2020pathvqa}: The PathVQA dataset consists of 4,998 images and 32,799 question-answer pairs generated from 1,670 pathology images collected from two pathology textbooks, and 3,328 pathology images collected from the PEIR digital library. The question are divided into seven categories.

\noindent\textbf{SLAKE} \cite{liu2021slake}: SLAKE contains 642 radiology images and more than 14,028 QA pairs. There are 2,603 triplets in English and 2,629 triplets in Chinese. The images are selected from three open source datasets \cite{kavur2021chaos,simpson2019large,wang2017chestx} and annotated by experienced experts.

\noindent\textbf{e-Ophta} \cite{decenciere2013teleophta}: The e-Ophta dataset comprises 47 images with the segmentation of hard exudates and 35 images without lesions.

\noindent\textbf{IDRiD} \cite{porwal2018indian}: The Indian Diabetic Retinopathy Image Dataset (IDRiD) contains color fundus images of the retina with a resolution of 4,288×2,848, whereby each of the 516 images contains disease severity levels as well as regions of four retinal biomarkers: \textit{microaneurysms, hemorrhages, hard exudates}, and \textit{soft exudates}.

\noindent\textbf{BACH} \cite{aresta2019bach}: The BreAst Cancer Histology Dataset (BACH) comprises over 400 labeled microscopy images. The images were annotated by two expert clinicians into four classes: \textit{normal, benign, in situ carcinoma}, and \textit{invasive carcinoma}.

\noindent\textbf{Tools} \cite{jin2018tool}: The Tool Detection and Operative Skill Assessment Dataset (Tools) consists of spatial bounding boxes of 2,532 frames across ten real-world laparoscopic surgical videos. With an average of 1.2 labels per frame, this dataset comprises 3,141 annotations on seven surgical tools.

We summarize information from commonly used datasets in table \ref{VQA_introduction}. In addition to these, the following datasets are also used for the medical VQA: Rad-ReStruct \cite{pellegrini2023rad}, RIS-VQA \cite{tascon2023localized}, INSEGCAT-VQA \cite{tascon2023localized}, OVQA \cite{huang2022ovqa}, EndoVis 2017 \cite{allan20192017} and EndoVis 2018 \cite{allan20202018}.
We show the results of different methods on various medical VQA datasets in Tables~\ref{VQA-RAD},~\ref{Pathvqa}, and~\ref{Slake}.

\begin{table*}[htbp]
\renewcommand\arraystretch{1.25}
\belowrulesep=0pt
\aboverulesep=0pt
  \centering
  \caption{Medical VQA results (accuracy) on VQA-RAD dataset.}
  \resizebox{\linewidth}{!}{
    \begin{tabular}{c|cccccc}
    \toprule
    \multirow{2}*{Model} & \multirow{2}*{Source} & \multirow{2}*{Highlights} & \multirow{2}*{Fusion} & \multicolumn{3}{c}{VQA-RAD} \\
    \cmidrule{5-7}          
    & & & & Close & Open & Overall \\
    \midrule
    \multirow{2}*{MMQ \cite{do2021multiple}} & \multirow{2}*{MICCAI 2021} & \multirow{2}*{Multiple Meta-model Quantifying} & SAN & 75.7 & 46.3 & 64 \\
    & & & BAN & 75.8 & 53.7 & 67 \\
           
    \multirow{3}*{CPRD \cite{liu2021contrastive}} & \multirow{3}*{MICCAI 2021} & Constrastive Learning & \multirow{3}*{BAN} & \multirow{3}*{80.4} & \multirow{3}*{61.1} & \multirow{3}*{72.7} \\
    & & Pre-train with extra data & & & &  \\
    & & Knowledge Distillation & & & &  \\

    MMBERT \cite{khare2021mmbert} & ISBI 2021 & Pre-train with extra data & Multi-head attention & 77.9  & 63.1  & 72 \\

    \multirow{3}*{VQAMix \cite{gong2022vqamix}} & \multirow{3}*{TMI 2022} & Cross-modal MixUp & SAN & 74±2.4 & 53.8±1.9 & 65.9±1.9 \\
    & & Learning with Missing Labels & \multirow{2}*{BAN} & \multirow{2}*{79.6±1.5} & \multirow{2}*{56.6±1.3} & \multirow{2}*{70.4±1.1} \\
    & & Learning with Conditional-mixed Labels & & & & \\

    \multirow{3}*{\cite{cong2022anomaly}} & \multirow{3}*{TMI 2022} & Weakly supervised anomaly localization & \multirow{2}*{BAN+Multiplication} & \multirow{2}*{82.05±0.66} & \multirow{2}*{64.48±2.01} & \multirow{2}*{75.07±1.00} \\
    & & Healthy image generation using GAN & & & &  \\
    & & Meta-Learning and multi-task learning & BAN+Residual & 80.39±1.20 & 60.89±2.26 & 72.65±0.86 \\
    
    CR \cite{liu2022medical} & TMI 2022 & Contrastive learning and conditional reasoning & BAN & 80.4 & 60.5 & 72.5 \\

    $MF^{2}-MVQA$ \cite{song2023mf} & ISBI 2023 & Multi-stage feature fusion & Multi-head attention & 80.1 & 63.7 & 73.6 \\

    \multirow{2}*{M2I2 \cite{li2023self}} & \multirow{2}*{ISBI 2023} & Masked image and language modeling & \multirow{2}*{Multi-head attention} & \multirow{2}*{83.5} & \multirow{2}*{66.5} & \multirow{2}*{76.8} \\
    & & Constrastive Learning & & & & \\

    \multirow{2}*{Q2ATransformer \cite{liu2023q2atransformer}} & \multirow{2}*{IPMI 2023} & Semi-open framework & \multirow{2}*{Concatenation} & \multirow{2}*{81.2} & \multirow{2}*{79.19} & \multirow{2}*{80.48} \\
    & & Learnable candidate answer embeddings & & & &  \\
    
    \multirow{2}*{DeBCF \cite{zhan2023debiasing}} & \multirow{2}*{MICCAI 2023} & DeBiasing Med-VQA model & \multirow{2}*{BAN} & \multirow{2}*{80.9±0.8} & \multirow{2}*{58.6±1.1} & \multirow{2}*{71.6±1.0} \\
    & & with CounterFactual training & & & & \\
    
    \multirow{2}*{MUMC \cite{li2023masked}} & \multirow{2}*{MICCAI 2023} & Masked image and text modeling & \multirow{2}*{Cross-attention} & \multirow{2}*{84.2} & \multirow{2}*{71.5} & \multirow{2}*{79.2} \\
    & & with Unimodal and Multimodal Contrastive losses & & & &  \\
          
    \multirow{3}*{hi-VQA \cite{pellegrini2023rad}} & \multirow{3}*{MICCAI 2023} & Structured radiology reporting benchmark & \multirow{3}*{Concatenation} & \multirow{3}*{-} & \multirow{3}*{-} & \multirow{3}*{76.3} \\
    & & Multi-question and multi-level tasks & & & &  \\
    & & Autoregressive formulation and consistent evaluation & & & &  \\
    \bottomrule
    \end{tabular}%
    }
  \label{VQA-RAD}%
\end{table*}%

\begin{table*}[htbp]
\renewcommand\arraystretch{1.25}
\belowrulesep=0pt
\aboverulesep=0pt
  \centering
  \caption{Medical VQA results (accuracy) on PathVQA dataset.}
  \resizebox{\linewidth}{!}{
    \begin{tabular}{c|cccccc}
    \toprule
    \multirow{2}*{Model} & \multirow{2}*{Source} & \multirow{2}*{Highlights} & \multirow{2}*{Fusion} & \multicolumn{3}{c}{PathVQA} \\
    \cmidrule{5-7}          
    & & & & Close & Open & Overall \\
    \midrule
    \multirow{2}*{MMQ \cite{do2021multiple}} & \multirow{2}*{MICCAI 2021} & \multirow{2}*{Multiple Meta-model Quantifying} & SAN & 82.7 & 11.2 & 47.1 \\
    & & & BAN & 84 & 13.4 & 48.8 \\

    \multirow{3}*{VQAMix \cite{gong2022vqamix}} & \multirow{3}*{TMI 2022} & Cross-modal MixUp & SAN & 84.4±0.2 & 12.1±0.5 & 48.4±0.2 \\
    & & Learning with Missing Labels & \multirow{2}*{BAN} & \multirow{2}*{83.5±0.2} & \multirow{2}*{13.4±0.6} & \multirow{2}*{48.6±0.3} \\
    & & Learning with Conditional-mixed Labels & & & & \\
    
    \multirow{2}*{Prefix T. Medical} & \multirow{2}*{MICCAI 2023} & Generation with prefix tuning & Attention & \multirow{2}*{87} & \multirow{2}*{40} & \multirow{2}*{63.6} \\
    LM \cite{van2023open} & & Parameter-efficient fine-tuning & mechanism & & & \\

    \multirow{2}*{MUMC \cite{li2023masked}} & \multirow{2}*{MICCAI 2023} & Masked image and text modeling & \multirow{2}*{Cross-attention} & \multirow{2}*{90.4} & \multirow{2}*{39.0} & \multirow{2}*{65.1} \\
    & & with Unimodal and Multimodal Contrastive losses & & & & \\

    \multirow{2}*{TraP-VQA \cite{naseem2022vision}} & \multirow{2}*{JBHI 2022} & High and low-level interactions & \multirow{2}*{Multi-head attention } & \multirow{2}*{93.57} & \multirow{2}*{37.72} & \multirow{2}*{64.82} \\
    & & Improve interpretability & & & & \\

    \multirow{2}*{M2I2 \cite{li2023self}} & \multirow{2}*{ISBI 2023} & Masked image and language modeling & \multirow{2}*{Multi-head attention} & \multirow{2}*{88} & \multirow{2}*{36.3} & \multirow{2}*{62.2} \\
    & & Constrastive Learning & & & & \\

    \multirow{2}*{Q2ATransformer \cite{liu2023q2atransformer}} & \multirow{2}*{IPMI 2023} & Semi-open framework & \multirow{2}*{Concatenation} & \multirow{2}*{88.85} & \multirow{2}*{54.85} & \multirow{2}*{74.61} \\
    & & Learnable candidate answer embeddings & & & & \\
    \bottomrule
    \end{tabular}%
    }
  \label{Pathvqa}%
\end{table*}%

\begin{table*}[htbp]
\renewcommand\arraystretch{1.25}
\belowrulesep=0pt
\aboverulesep=0pt
  \centering
  \caption{Medical VQA results (accuracy) on SLAKE dataset. ``Eng'' means that the model only uses the English question-answer pairs on SLAKE dataset.}
  \resizebox{\linewidth}{!}{
    \begin{tabular}{c|cccccc}
    \toprule   
    \multirow{2}*{Model} & \multirow{2}*{Source} & \multirow{2}*{Highlights} & \multirow{2}*{Fusion} & \multicolumn{3}{c}{SLAKE} \\
    \cmidrule(lr){5-7} 
    &  &  &  & Close &  Open & Overall \\
    \midrule
    \multirow{3}*{CPRD(Eng) \cite{liu2021contrastive}} &  \multirow{3}*{MICCAI 2021} & Constrastive Learning & \multirow{3}*{BAN} &  \multirow{3}*{83.4} & \multirow{3}*{81.2} & \multirow{3}*{82.1} \\
    & & Pre-train with extra data & & & & \\
    & & Knowledge Distillation & & & & \\
    CR(Eng) \cite{liu2022medical} & TMI 2022 & Contrastive learning and conditional reasoning & BAN & 84.1 & 80.5 & 81.9 \\
    \multirow{2}*{DeBCF(Eng) \cite{zhan2023debiasing}} & \multirow{2}*{MICCAI 2023} & DeBiasing Med-VQA model & \multirow{2}*{BAN} & \multirow{2}*{84.9±0.7} & \multirow{2}*{80.8±0.9} & \multirow{2}*{82.6±0.9} \\
    & & with CounterFactual training & & & & \\
    \multirow{2}*{Prefix T. Medical} & \multirow{2}*{MICCAI 2023} & Generation with prefix tuning & Attention & \multirow{2}*{82.1} & \multirow{2}*{84.3} & \multirow{2}*{83.3} \\
    LM(All) \cite{van2023open}& & Parameter-efficient fine-tuning & mechanism & & & \\
    \multirow{2}*{MUMC(All) \cite{li2023masked}} & \multirow{2}*{MICCAI 2023} & Masked image and text modeling & \multirow{2}*{Cross-attention} & \multirow{2}*{-} & \multirow{2}*{-} & \multirow{2}*{84.9} \\
    & & with Unimodal and Multimodal Contrastive losses & & & & \\
    \multirow{2}*{M2I2(All) \cite{li2023self}} & \multirow{2}*{ISBI 2023} & Masked image and language modeling & \multirow{2}*{Multi-head attention} & \multirow{2}*{91.1} & \multirow{2}*{74.7} & \multirow{2}*{81.2} \\
    & & Constrastive Learning & & & & \\
    \bottomrule
    \end{tabular}%
    }
  \label{Slake}%
\end{table*}%

\subsubsection{Metrics}

\noindent\textbf{Consistency C1} measures the ratio of correctly answered sub-questions when the main question is correctly answered. Consistency C2 measures the ratio of correctly answered main questions when all corresponding sub-questions are correctly answered.

\noindent\textbf{Accuracy} evaluates the percentage of correct responses of the model for all questions, \ie the number of correct responses divided by the total number of questions.

\noindent\textbf{Micro-accuracy} is the number of correct predictions divided by the total number of predictions, calculated by aggregating the results of all questions.

\noindent\textbf{Macro-accuracy} calculates the average of the accuracy of the predictions for all problems.

\noindent\textbf{Top-k accuracy} is the proportion of the model's predictions in which the correct answer appears in the top-k predictions.

\noindent\textbf{Median accuracy} calculates the median accuracy of the predictions for all questions.

\noindent\textbf{Mean reciprocal rank (MRR)} is the mean of the reciprocal of the proportion of correct answers that the model ranks first when predicting the answer.

\subsection{Datasets in Medical Multimodal Diagnosis and Prognosis}

The following datasets are used for the medical image diagnosis and prognosis task: Musculoskeletal Radiographs (MURA) \cite{rajpurkar2017mura}, Montgomery \cite{jaeger2014two}, Shenzhen Dataset \cite{jaeger2014two}, IDRiD \cite{porwal2018indian}, CheXpert \cite{chexpert_irvin2019chexpert}, ChestX-ray14 \cite{wang2017chestx}, China Consortium of Chest CT Image Investigation (CC-CCII) \cite{zhang2020clinically}, NSCLC \cite{kiser2020data}, \cite{elmore2015diagnostic}, PLCO \cite{gohagan2000prostate}, QaTa-COV19 \cite{degerli2022osegnet}, IU X-ray \cite{iu_xray_demner2016preparing}, LIDC-IDRI \cite{armato2011lung}, MIMIC-CXR \cite{MIMIC-CXR_johnson2019mimic}, Chest X-ray-8 \cite{wang2017chestx}, NSCLC-Radiomic \cite{clark2013cancer}. 

\subsection{Datasets in Text-guided Medical Image Segmentation}

The following datasets are used for the medical image segmentation: IDRiD \cite{porwal2018indian}, DDR \cite{li2019diagnostic}, QaTa-COV19 \cite{degerli2022osegnet}, MosMedData+ \cite{morozov2020mosmeddata}, RadImageNet \cite{radford2021learning}, Medical Segmentation Decathlon (MSD) \cite{antonelli2022medical}, Full Adult Fly Brain (FAFB) \cite{schlegel2021automatic}, MitoEM \cite{wei2020mitoem}, FIB-25 \cite{takemura2017connectome}, Kasthuri15 \cite{kasthuri2015saturated}, SABSCT \cite{landman2015miccai}, CHAOS \cite{kavur2021chaos}, MS-CMR \cite{zhuang2022cardiac}, RIGA+ \cite{hu2022domain}, Breast Ultrasound Images dataset (BUSI) \cite{al2020dataset}, COVID-19 Radiography Database (COVID-QU-Ex) \cite{chowdhury2020can,rahman2021exploring}, MoNuSeg \cite{kumar2019multi}, Glas \cite{sirinukunwattana2017gland}, LUAD-HistoSeg \cite{han2022multi}, BCSS \cite{amgad2019structured}, HC18 \cite{van2018automated}, Multi-Modality Whole Heart Segmentation (MMWHS) Challenge 2017 \cite{zhuang2016multi,zhuang2013challenges}, REFUGE2 \cite{fang2022refuge2}, PALM \cite{fang2023palm}, BTCV \cite{landman2015miccai}.

\subsection{Datasets in Medical Image-Text Retrieval}
The following datasets are used for the medical image-text retrieval: MIMIC-CXR~\cite{MIMIC-CXR_johnson2019mimic}, CheXpert~\cite{chexpert_irvin2019chexpert}, PTB-XL~\cite{ptb-xl_wagner2020ptb}, OpenI~\cite{iu_xray_demner2016preparing}, ROCO~\cite{roco_pelka2018radiology}, BIMCV-R~\cite{chen2024bimcv}, 3D-MIR~\cite{abacha20233d}.

\subsubsection{Metrics}
\noindent\textbf{Mean Average Precision (mAP)} evaluates the precision of retrieval across different recall levels. It is calculated by averaging the precision scores at each relevant item retrieved and is often used to assess the overall performance of a retrieval model.

\noindent\textbf{Precision at K (P@K)} measures the proportion of relevant items within the top K retrieved results. It focuses on the relevance of the most highly ranked items, making it particularly useful in scenarios where users are most interested in the top results.

\noindent\textbf{Recall at K (R@K)} calculates the proportion of all relevant items that are retrieved within the top K results. It emphasizes the ability of the model to find as many relevant items as possible in the top K positions.

\noindent\textbf{Mean Reciprocal Rank (MRR)} focuses on the position of the first relevant item in the retrieval results. It is calculated by averaging the reciprocal ranks of the first relevant result for each query, providing insight into how quickly relevant information is found.

\end{appendices}


\end{document}